\documentclass[10pt,twocolumn,letterpaper]{article}

\usepackage[pagenumbers]{cvpr} 

\usepackage{graphicx}

\usepackage{graphicx}

\usepackage{amsmath,amssymb,amsthm}

\usepackage{adjustbox}
\usepackage{url}
\usepackage{longtable}
\definecolor{cvprblue}{rgb}{0.21,0.49,0.74}
\usepackage[pagebackref,breaklinks,colorlinks,allcolors=cvprblue]{hyperref}

\def\confName{CVPR}
\def\confYear{2027}

\title{From Frozen Features to Gaussian Token Spaces:\\ Gradient-Free Kernel Prediction for Earth Observation}

\author{First Author\\
Institution1\\
Institution1 address\\
{\tt\small firstauthor@i1.org}
\and
Second Author\\
Institution2\\
First line of institution2 address\\
{\tt\small secondauthor@i2.org}
}

\author{
  Dominik J. M\"uhlematter$^{1,2*}$ \quad
  Philip Janz$^{1,2}$ \quad
  Xue Xia$^{1}$\\
  Konrad Schindler$^{2}$\quad 
  Helge Aasen$^{1}$ \quad 
  Mehmet Ozgur Turkoglu$^{1*\dagger}$ \\[2ex]
  $^{1}$Agroscope, Earth Observation of Agroecosystems, Zurich, Switzerland \\
  $^{2}$ETH Zurich, Switzerland \\[0.5ex]
  {\small $^{*}$equal contribution \quad $^{\dagger}$corresponding author} \\
  \small\texttt{moturkoglu@gmail.com}
}
\begin{document}
\maketitle
\begin{abstract}
To make use of Earth observation (EO) foundation models, users must train task-specific decoders, and often also fine-tune the foundational encoder for their task. This adds computational cost and can lead to overfitting when labelled examples are scarce. We propose an alternative way to employ pretrained foundation models: we show that one can, without any gradient-based optimisation, reorganise the geometry of the pretrained feature space such that it becomes decodable with elementary predictors. Using ideas from self-supervised feature learning, we warp the feature space to give it isotropic Gaussian structure, and show that in the transformed space accurate predictions can be made with simple kernel-based prediction. The principle generalizes across different EO foundation models and makes it easy to integrate additional features such as DINOv3, as well as classical spectral indices. 
Our approach is computationally very cheap and takes only a few minutes to set up. Across multiple EO benchmarks, its performance is comparable to the state of the art (sometimes even better). Moreover, it remains effective in data-scarce and few-shot settings.

\end{abstract}

\section{Introduction}
\label{sec:introduction}

\begin{figure*}[th]
    \centering
    \includegraphics[width=1.0\textwidth]{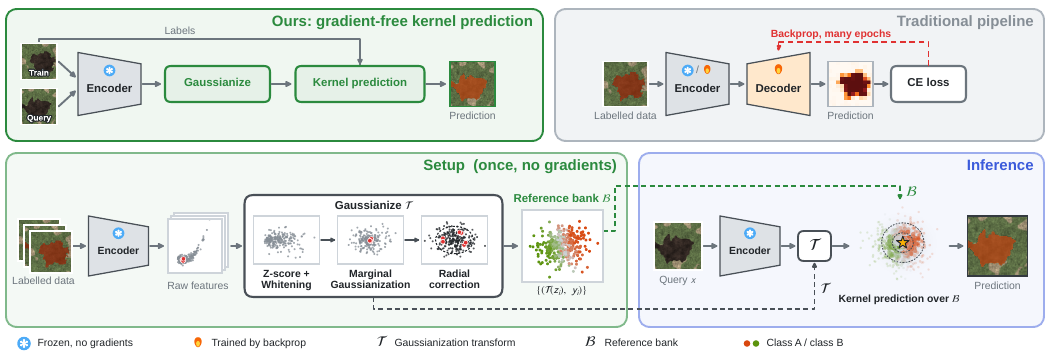}
   \caption{\textbf{Method overview.} 
    Features are extracted from labelled training images with a frozen encoder; mapped to an isotropic Gaussian distribution by a post-hoc transformation $\mathcal{T}$ consisting of $z$-scoring, whitening, marginal Gaussianization, and radial correction; and stored in a reference bank $\mathcal{B}$. At inference, a query image undergoes the same encoding and transformation. Predictions are obtained through kernel-weighted aggregation of the labelled entries in $\mathcal{B}$. No decoder is trained and no gradients are computed.}
    \label{fig/method}
\end{figure*}

Foundation models, including those developed for Earth observation (e.g.~\cite{fuller2023croma,jakubik2025terramind}), are pretrained on large datasets at substantial computational cost. The goal is to learn generic representations that are suitable for different downstream analysis tasks. If pretraining achieves that goal, features extracted with the learned, frozen encoder should enable inference across diverse tasks with relatively simple predictors, as demonstrated in computer vision~\cite{oquab2023dinov2} and natural language processing applications~\cite{reimers2019sentence}.

In EO, however, downstream adaptation often involves training a high-capacity decoder such as UPerNet~\cite{xiao2018unified}, and in many cases also fine-tuning the entire encoder. Consequently, a significant volume of training data and compute is required, thus losing a central advantage and a main motivation for using foundation models. Note that also pre-computed embeddings~\cite{brown2025alphaearth,feng2026tessera} are typically used with high-capacity decoders~\cite{ronneberger2015u}, moreover, it is not yet clear how generally they are applicable~\cite{herzog2026olmoearth}. 
A further limitation of the standard procedure is its reliance on labelled training data for the task head and for encoder fine-tuning. If only few labelled data are available, performance can drop sharply, as even relatively lightweight decoders will tend to overfit (\cf Sec.~\ref{sec:exp}).
The cause of both problems is the same: the feature encoding is not regarded as a foundational representation of the knowledge needed for a range of prediction tasks, but merely as a ``useful starting point'' that requires extensive further training to build a predictor.

This motivates our central question: \textit{Can we make better use of pretrained EO representations, without relying on expensive downstream adaptation?} Our hypothesis is that the features already contain the information needed to solve many tasks; but that, in the original feature space, this information may not be readily accessible.

In particular, the fact that the information is \emph{present} does not necessarily imply that the geometric layout of the space makes it \emph{accessible} to shallow, low-capacity probes. Pretrained features may be highly anisotropic, have spatially varying scatter, or exhibit ``hubness'' (i.e., a small set of data points that are nearest neighbours for a large fraction of all others~\cite{radovanovic2010hubs}). These properties hide information from predictors based on simple (e.g., Euclidean or Manhattan) distances.
Put differently, the geometry of the feature space does not necessarily reflect the relations relevant for downstream use, rather it is shaped by the pretraining data, the training objective, and the optimisation dynamics. Prior work in few-shot image classification has shown that even elementary forms of feature normalisation can improve nearest-neighbour prediction~\cite{wang2019simpleshot}; supporting our argument that the information can be present in the feature space, but distorted in a way that makes it inaccessible, thus requiring high-capacity decoders (which can be seen as complex mapping functions to "undistort" the space).

Our approach is to keep the feature representation ``frozen'' (in the sense of not applying gradient-based learning to it), but to deterministically warp it in a way that exposes relevant relations, without an elaborate, learnable decoder. If such a warping can be found, then different tasks should become solvable with lightweight "decoders" like kernel regression, or even $k$-nearest neighbours.

Our work is inspired by variational autoencoders (VAEs,~\cite{kingma2013auto}), where regularisation towards a benign prior, typically an isotropic Gaussian, structures the latent space such that it supports generation. Also, for self-supervised representation learning, LeJEPA~\cite{balestriero2025lejepa} offers a theoretical motivation for an isotropic Gaussian embedding space (under sensible assumptions) and encourages that distribution through a suitable regulariser. 
Building on these ideas, we ask whether a similar structure can be imposed post-hoc to make pretrained EO features more readable for cheap, distance-based predictors. Empirically, we find that this is indeed the case.
Our contributions are:
\begin{itemize}
    \item \textbf{An alternative view on adapting and decoding pretrained features.}
    We develop a geometrically motivated adaptation scheme that combines $z$-score normalisation, PCA whitening, inverse-CDF Gaussianisation~\cite{laparra2011rbig}, and radial correction~\cite{lyu2009radial} to achieve a predictor-friendly (isotropic, Gaussian) feature distribution. The adapted feature space supports distance-based prediction, without encoder
    fine-tuning or gradient-based decoder training.

    \item \textbf{A simple and label-efficient baseline for EO foundation models.}
    Using off-the-shelf, frozen encoders, our method reaches competitive performance on popular EO benchmarks; in some cases even setting a new state-of-the-art. Furthermore, it maintains good performance also in data-scarce and few-shot settings, where conventional, training-based adaptation degrades.

\end{itemize}
\section{Method}
\label{sec:method}

\begin{figure}[th]
    \centering
    \includegraphics[width=\columnwidth]{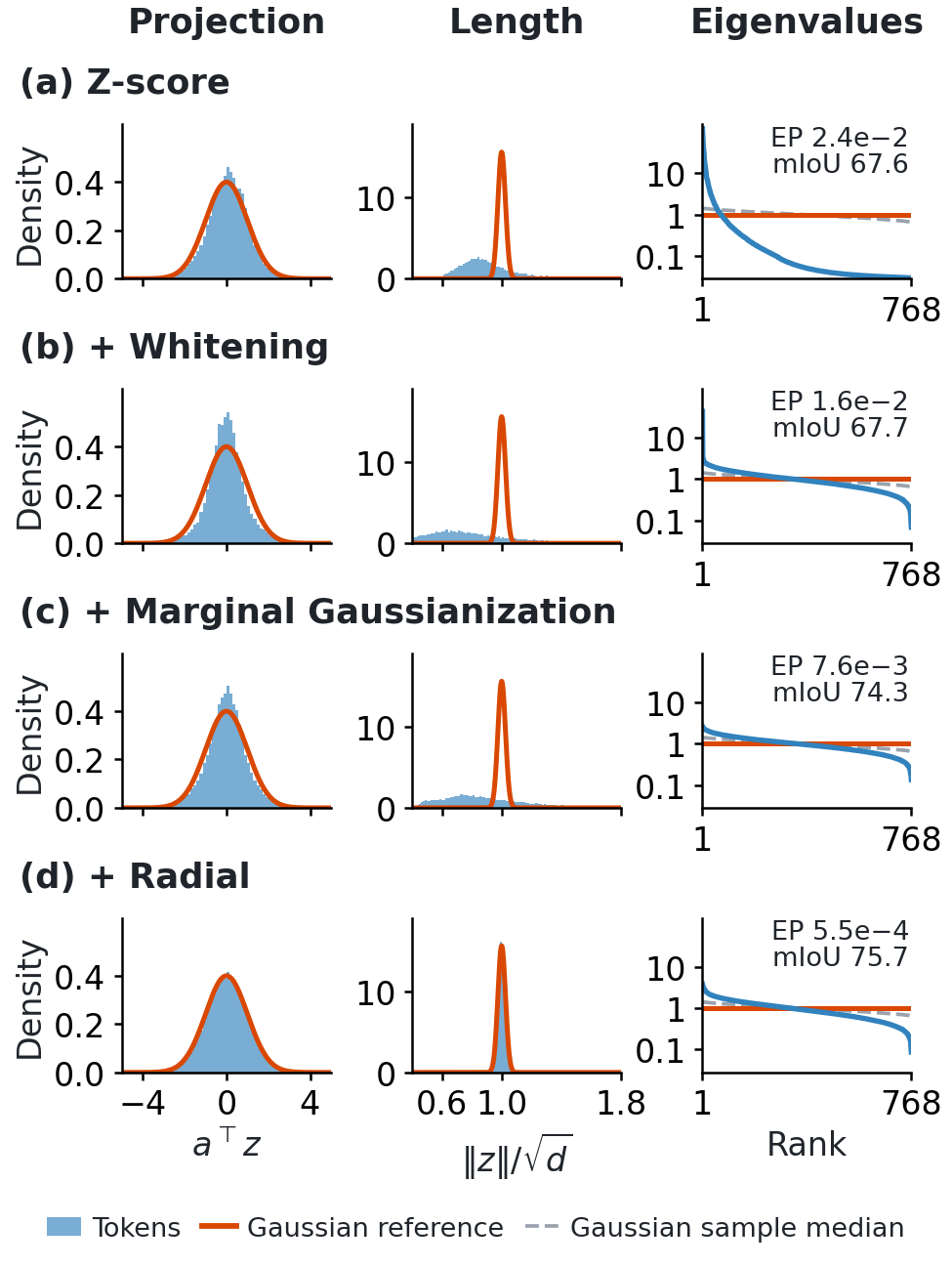}
    \caption{\textbf{Effect of post-hoc isotropic Gaussianization on MADOS using OlmoEarth v1.3 Base and Gaussian kernel prediction.} Columns show random 1D projections, normalized lengths, and covariance eigenvalues. EP is the projection-based Epps--Pulley discrepancy (lower is better)~\cite{balestriero2025lejepa}; mIoU reports full-pipeline test segmentation performance.}
    \label{fig:token_space_analysis}
\end{figure}
We predict image and pixel labels from annotated references using a frozen encoder. 
We calibrate the feature space before predicting labels directly from a labelled reference bank using simple distance-based predictors, without training a task-specific decoder. For segmentation, a pixel branch supplies spatial detail, and the resulting score maps are fused and refined (Fig.~\ref{fig/method}). No downstream gradient-based optimization is required.

\paragraph{Feature groups and reference-based prediction.}
We form a visual group from the frozen encoder's final representation,
concatenating descriptors across input views where applicable
(e.g., RGB and a near-infrared false-color composite).
A complementary spectral group contains the mean, standard deviation,
and average gradient magnitude of bands and available spectral indices
(NDVI, NDWI, NDBI, NBR, and BSI).
These statistics summarize each token's image footprint for segmentation
and the whole image for classification.
For time series, descriptors are concatenated across dates within
each group.

For each group, we fit a post-hoc feature transformation on training
descriptors and apply it identically to references and queries.
The transformation is described below.
We construct a reference bank $\mathcal{B}$ containing patch-token
descriptors for segmentation or image-level descriptors for classification.
Each reference $a\in\mathcal{B}$ pairs a transformed descriptor
$z_a\in\mathbb{R}^d$ with a class vector $y_a$.
For segmentation, $y_a$ gives the class proportions among labeled pixels
in the corresponding footprint; training tokens containing no labeled
pixels are omitted.
For classification, $y_a$ is the one-hot image label.

Given a transformed query descriptor $z$, Gaussian-kernel
regression~\cite{nadaraya1964estimating,watson1964smooth} predicts
\begin{equation}
S_c(z)=
\frac{\sum_{a\in\mathcal{B}}\kappa_h(z,z_a)y_{a,c}}
     {\sum_{a\in\mathcal{B}}\kappa_h(z,z_a)},
\label{eq:method_kernel}
\end{equation}
where $\kappa_h(z,z_a)=\exp(-\|z-z_a\|_2^2/(2h^2))$ and the bandwidth
$h$ controls the locality of the vote.
We also evaluate cosine kNN on normalized descriptors.
Its neighborhood size and weighting rule (uniform or exponential
in cosine similarity) are selected separately for each group on
validation data, as is the Gaussian-kernel bandwidth.

\paragraph{Descriptor geometry for kernel prediction.}
Following the analysis of LeJEPA~\cite{balestriero2025lejepa}, we examine how descriptor
density affects Gaussian kernel prediction. Let $p$ denote the descriptor density and $\eta_c(z)=\mathbb{E}[y_c\mid z]$ the conditional target. Under standard smoothness assumptions, the population counterpart of Eq.~\eqref{eq:method_kernel} has small-bandwidth bias
\begin{equation}
\!\!\operatorname{bias}_c(z)\!=\!\tfrac{h^2}{2}\Delta\eta_c(z)\hspace{-.7pt}+\hspace{-.7pt}h^2\nabla\eta_c(z)^{\!\top\!}\nabla\hspace{-.5pt}\log p(z)\hspace{-.7pt}+\hspace{-.7pt}o(h^{\!2}).
\label{eq:method_bias}
\end{equation}
The first term reflects target curvature; the second captures how
uneven reference density shifts predictions toward denser regions. For bounded target derivatives, the leading squared bias,
averaged over $z\sim p$, admits an upper bound that increases
linearly with the Fisher information
$J(p)\!=\!\mathbb{E}_{z\sim p}\|\nabla\log p(z)\|_2^2$. Among smooth densities with fixed total variance, isotropic Gaussians minimize $J(p)$.

These results motivate an isotropic Gaussian target. We investigate whether transforming frozen descriptors toward this target makes their task-relevant information more accessible to reference-based predictors. Further details are provided in
Appendix~\ref{app:theoretical_details}.

\paragraph{Post-hoc isotropic Gaussianization.}
We transform each visual and spectral feature group toward $\mathcal{N}(0,I_d)$ using training-set statistics. Figure~\ref{fig:token_space_analysis} illustrates the four stages through
their effects on feature projections, descriptor lengths, and covariance eigenvalues. All transformations are fitted on training descriptors and applied unchanged to queries.

\noindent
\textit{(a) Z-score.}
We subtract the training mean of each feature coordinate and divide by its standard deviation. This places coordinates on a common scale, but leaves correlations between them.

\noindent
\textit{(b) Whitening.}
We apply PCA, retain $d$ components, and rescale each to unit variance. The resulting descriptor $u\in\mathbb{R}^d$ has decorrelated coordinates. As the figure illustrates, whitening can nevertheless leave non-Gaussian distributions and substantial variation in descriptor lengths.

\noindent
\textit{(c) Marginal Gaussianization.}
Following~\cite{laparra2011rbig}, we map each coordinate to the Gaussian quantile corresponding to its position in the training distribution:
\begin{equation}
u_j\leftarrow
\Phi^{-1}\!\big(
\operatorname{clip}_{\epsilon}(\widehat F_j(u_j))
\big),
\qquad j=1,\ldots,d,
\label{eq:method_marginal}
\end{equation}
where $\widehat F_j$ is the empirical training CDF and $\Phi$ the standard normal CDF. Clipping avoids probabilities near zero and one. We repeat this operation for $K$ rounds, each followed by PCA re-whitening to remove correlations introduced by the nonlinear maps and define the axes for the next round.

\textit{(d) Radial correction.}
Gaussian-looking projections can still coexist with a broad distribution of lengths, as shown in Fig.~\ref{fig:token_space_analysis}(c). This matters for Euclidean retrieval: under Gaussian queries, shorter references have a pairwise advantage, consistent with the centrality mechanism underlying hubness \cite{radovanovic2010hubs}. We therefore apply radial Gaussianization~\cite{lyu2009radial}, matching descriptor lengths to those of a standard Gaussian:
\begin{equation}
z=\tfrac{u}{\|u\|_2}\,
H_d^{-1}\!\big(
\operatorname{clip}_{\epsilon}(\widehat F_R(\|u\|_2))
\big),
\label{eq:method_radial}
\end{equation}
where $\widehat F_R$ is the training norm CDF and $H_d$ the CDF of $\chi_d$, the Gaussian norm distribution in $d$ dimensions. The map adjusts lengths while preserving directions and cosine similarities. These finite, empirically fitted transformations approximate the target geometry without guaranteeing joint Gaussianity.

\paragraph{Pixel branch.}
Token predictions describe class proportions within a patch but do not locate its classes. The pixel branch recovers this spatial detail through two-stage matching (Fig.~\ref{fig/method}). First, each query patch retrieves reference patches by cosine similarity between their transformed visual descriptors. The labeled pixels in these reference patches form a candidate bank for all pixels in the query patch. Each query and reference pixel is represented by its band values, available spectral indices, and per-band means and standard deviations in a small neighborhood. We standardize and apply regularized PCA whitening to these descriptors, followed by radial correction, using transformations fitted on training pixels. No marginal Gaussianization is applied. Each query pixel then finds its nearest reference pixels by cosine similarity within the candidate bank, regardless of their positions inside the reference patches.
Their class labels vote with exponential similarity weights, producing a class-score vector for the query pixel.

\paragraph{Late fusion.}
We combine the visual and spectral token predictions with the pixel-level prediction after correcting for class imbalance. We apply prior correction, equivalent to post-hoc logit adjustment~\cite{menon2021longtail}:
$\widetilde S_c^g\propto S_c^g/\pi_c^{\alpha_g}$, with normalization over classes. Here, $\pi_c$ is the average class-fraction vector of the reference tokens. The exponent $\alpha_g$ controls the correction strength: $\alpha_g=0$ leaves scores unchanged, while positive values increase the relative contribution of rare classes. Each token branch's exponent is selected with its predictor settings on validation data; the pixel branch inherits the exponent of the visual retrieval branch.

We upsample each corrected token map to image resolution, interpolating its class-score channels separately before fusion. Each pixel therefore receives a class-score vector from each of the visual, spectral, and pixel branches. Denoting these aligned maps by $A_c^g(x)$, we compute
\begin{equation}
P_c(x)\propto\prod_g A_c^g(x)^{w_g},
\qquad
w_g\geq0,\quad \sum_g w_g=1,
\label{eq:method_fusion}
\end{equation}
and normalize over classes. The validation-selected weights are shared across pixels and control the contribution of each branch.

For spatial refinement, we apply SLIC~\cite{achanta2012slic} to standardized image-band features. SLIC groups pixels using both spectral similarity and spatial proximity, producing connected superpixels independently of the patch-token grid.
Let $\mathcal R(x)$ denote the superpixel containing pixel $x$ and
$\overline P_c(\mathcal R(x))$ its average fused score for class $c$.
We refine the scores as
\begin{equation}
P_c^{\mathrm{final}}(x)
=\beta P_c(x)
+(1-\beta)\overline P_c(\mathcal R(x)),
\label{eq:method_refinement}
\end{equation}
where $\beta$ is selected on validation data.
The final label is $\arg\max_c P_c^{\mathrm{final}}(x)$. Classification uses the same prior correction and geometric fusion on image-level scores, with image-label frequencies as the prior, and omits pixel prediction and spatial refinement.
For further implementation details, see Appendix~\ref{app:method_implementation_details}.
\section{Experiments}\label{sec:exp}
EO foundation models are commonly evaluated using trained
task-specific heads on frozen backbones~\cite{marsocci2026pangaea}.
We test whether our method offers a competitive alternative
without downstream gradient-based optimization, across EO tasks,
backbones, and label budgets.

\paragraph{Baselines.}
For segmentation, we compare against a \textit{Light} decoder and
\textit{UPerNet}~\cite{xiao2018unified} (61k vs. 47M parameters with CROMA-L on MADOS), a widely used decoder adopted
by PANGAEA~\cite{marsocci2026pangaea}. \textit{UPerNet} combines pyramid pooling and feature-pyramid fusion
of multilevel features.
Our \textit{Light} decoder applies a single trainable $1\times1$ convolution
to spatially aligned, concatenated encoder features from selected
layers and, where applicable, dates, then bilinearly upsamples
the logits.
For classification, we compare against linear probes on frozen
image representations.
We also implement multiclass versions of the training-free
segmentation methods FROST~\cite{park2026frost} and
INSID3~\cite{cuttano2026insid3}.

\paragraph{Datasets and evaluation.}
We evaluate on six benchmarks spanning optical, temporal,
and SAR imagery, and both dense prediction and scene classification.
MADOS and HLS Burn Scars assess optical semantic segmentation
of marine debris and sea-surface features, and burned areas,
respectively.
PASTIS evaluates crop segmentation from multispectral time series,
xView2 evaluates building damage segmentation from high-resolution
pre- and post-disaster image pairs, and Sen1Floods11 S1 assesses
flood segmentation from Sentinel-1 SAR imagery.
For these dense prediction tasks, we follow PANGAEA data
preparation and splits~\cite{marsocci2026pangaea} and report
mean intersection-over-union (mIoU).
For EuroSAT scene classification~\cite{helber2019eurosat}, we report top-1 accuracy on spatial splits~\cite{robinson2026eurosat}.

\paragraph{Backbones.}
We evaluate DINOv3 ViT-L/16 \cite{simeoni2025dinov3}, CROMA-L \cite{fuller2023croma}, DOFA ViT-B \cite{xiong2024dofa},
Scale-MAE ViT-L ~\cite{reed2023scalemae}, TerraMind 1.0 Base \cite{jakubik2025terramind}, and OlmoEarth v1.3 Base \cite{tseng2026olmoearthv13}.
CROMA uses its optical or SAR branch according to the input
modality.
All backbone parameters remain frozen during downstream evaluation.

\paragraph{Experimental protocol.}
We consider full-label settings and reduced-label settings
spanning data-scarce and few-shot regimes.
The 10\%/1\% label budgets correspond to 48/4 training entries
for HLS Burn Scars, 143/14 for MADOS, and
145/14 for PASTIS.
For xView2, we use 1\%/0.1\% budgets, yielding 133/13 entries
and comparable subset sizes to MADOS and PASTIS.
Training data populate reference banks and determine feature
transformations.
Validation data select between the kernel-based classifier
and cosine kNN, together with their hyperparameters, fusion
weights, and spatial refinement settings.
Trained decoder and classifier baselines use PANGAEA-based
training protocols.
Further results and experimental details are provided in
Appendices~\ref{app:additional_results}
and~\ref{app:experimental_setup}.

\subsection{Main Results}

\begin{table*}[t]
\centering
\setlength{\tabcolsep}{3pt}
\renewcommand{\arraystretch}{1.08}
\begin{tabular}{@{}l
    ccc @{\hspace{5pt}}|@{\hspace{5pt}}
    ccc @{\hspace{5pt}}|@{\hspace{5pt}}
    ccc @{\hspace{5pt}}|@{\hspace{5pt}}
    ccc@{}}
\toprule
& \multicolumn{3}{c}{HLS Burn Scars}
& \multicolumn{3}{c}{MADOS}
& \multicolumn{3}{c}{PASTIS}
& \multicolumn{3}{c}{xView2} \\
\cmidrule(lr){2-4}
\cmidrule(lr){5-7}
\cmidrule(lr){8-10}
\cmidrule(lr){11-13}
FROST \cite{park2026frost}
& \multicolumn{3}{c}{66.7}
& \multicolumn{3}{c}{39.4}
& \multicolumn{3}{c}{15.6}
& \multicolumn{3}{c}{44.3} \\
INSID3 \cite{cuttano2026insid3}
& \multicolumn{3}{c}{49.5}
& \multicolumn{3}{c}{36.6}
& \multicolumn{3}{c}{12.9}
& \multicolumn{3}{c}{32.6} \\
\midrule
Backbone
& Light & UPerNet & Ours
& Light & UPerNet & Ours
& Light & UPerNet & Ours
& Light & UPerNet & Ours \\
\midrule
CROMA optical
& 80.6 & 82.0 & \textbf{84.7}
& 56.8 & 67.6 & \textbf{69.3}
& 34.8 & 32.1 & \textbf{38.8}
& 37.9 & \textbf{53.0} & 44.4 \\
DOFA
& 79.7 & 79.0 & \textbf{84.2}
& 59.8 & 59.6 & \textbf{75.5}
& 26.4 & 24.5 & \textbf{35.5}
& 49.6 & \textbf{59.7} & 47.7 \\
Scale-MAE
& 74.3 & 76.7 & \textbf{85.2}
& 57.4 & 57.3 & \textbf{65.2}
& 21.8 & 11.7 & \textbf{35.0}
& 51.8 & \textbf{60.4} & 44.7 \\
TerraMind 1.0 Base
& 79.9 & 82.5 & \textbf{84.7}
& 60.0 & \textbf{72.0} & 68.7
& 37.0 & 29.0 & \textbf{37.3}
& 49.5 & \textbf{59.8} & 47.2 \\
OlmoEarth v1.3 Base
& 85.7 & \textbf{86.9} & 84.6
& 74.5 & \textbf{79.5} & 74.1
& 41.4 & 39.0 & \textbf{44.4}
& 40.2 & \textbf{53.2} & 44.9 \\
DINOv3 ViT-L/16
& 85.0 & \textbf{87.4} & 85.4
& 73.1 & 73.8 & \textbf{79.2}
& 30.2 & 28.6 & \textbf{35.8}
& 54.0 & \textbf{62.1} & 55.5 \\
\midrule
Mean
& 80.9 & 82.4 & \textbf{84.8}
& 63.6 & 68.3 & \textbf{72.0}
& 31.9 & 27.5 & \textbf{37.8}
& 47.2 & \textbf{58.0} & 47.4 \\
\bottomrule
\end{tabular}
\caption{
Test mIoU (\%) on HLS Burn Scars, MADOS, PASTIS, and xView2.
\textbf{Bold} indicates the best available method for each
backbone and dataset, and the best mean within each dataset.
}
\label{tab:backbone-comparison-four-datasets}
\end{table*}
\paragraph{Optical segmentation.} Tab.~\ref{tab:backbone-comparison-four-datasets} compares our
method with trained \textit{Light} and \textit{UPerNet} decoders.
Across six frozen backbones, our method achieves average mIoU
of 84.8\% on HLS Burn Scars, 72.0\% on MADOS, and 37.8\%
on PASTIS, exceeding the stronger decoder baseline by
2.4, 3.7, and 5.9 pp, respectively.
The results demonstrate competitive performance across optical segmentation
and temporal crop segmentation,without having to train a task-specific
decoder.
With DINOv3 backbone, our method also outperforms FROST and INSID3
on all four benchmarks.

On xView2, our method remains competitive but performs below \textit{UPerNet}.
Its greater decoding capacity and learned multiscale fusion
may help \textit{UPerNet} adapt frozen features to the spatial resolution
and object scales of high-resolution imagery, particularly
when recovering fine building boundaries.
This suggests that learned spatial adaptation may be more
beneficial in this setting.


\begin{table}[t]
\centering
\small
\setlength{\tabcolsep}{3pt}
\renewcommand{\arraystretch}{1.08}
\begin{tabular*}{\columnwidth}{@{\extracolsep{\fill}}lccc@{}}
\toprule
Backbone & Light & UPerNet & Ours \\
\midrule
CROMA SAR
& 78.4 & 79.2 & \textbf{80.2} \\
DOFA
& 74.2 & 78.8 & \textbf{80.1} \\
Scale-MAE
& 77.6 & \textbf{79.5} & 78.4 \\
TerraMind 1.0 Base
& 78.1 & 81.7 & \textbf{81.7} \\
OlmoEarth v1.3 Base
& 80.5 & \textbf{81.0} & 80.4 \\
DINOv3 ViT-L/16
& 76.5 & \textbf{80.3} & 79.7 \\
\midrule
Mean
& 77.6 & \textbf{80.1} & \textbf{80.1} \\
\bottomrule
\end{tabular*}
\caption{
SAR-only flood segmentation on Sen1Floods11:
test mIoU (\%). All methods receive Sentinel-1 as input; Scale-MAE and DINOv3 use SAR-derived three-channel representations.
\textbf{Bold} indicates the best method per backbone, mean denotes the average over all six backbones.
}
\label{tab:sar-comparison}
\end{table}
\paragraph{SAR segmentation.} Tab.~\ref{tab:sar-comparison} reports SAR-only flood segmentation
on Sen1Floods11.
Our method achieves 80.1\% mean mIoU, closely matching \textit{UPerNet}
at 80.1\% and exceeding \textit{Light} by 2.5 percentage points without
training a task-specific decoder.
It performs best with CROMA SAR, DOFA, and TerraMind, while the advantage of \textit{UPerNet} with optical-pretrained Scale-MAE
and DINOv3 suggests that learned decoding may help compensate
for the modality mismatch when using frozen features.

\begin{table}[t]
\centering
\small
\setlength{\tabcolsep}{3pt}
\renewcommand{\arraystretch}{1.08}
\begin{tabular*}{\columnwidth}{@{\extracolsep{\fill}}lcc@{}}
\toprule
Backbone & Linear & Ours \\
\midrule
CROMA optical
& 92.2 & \textbf{95.4} \\
DOFA
& \textbf{93.5} & 93.4 \\
Scale-MAE
& 93.3 & \textbf{94.2} \\
TerraMind 1.0 Base
& 93.7 & \textbf{95.8} \\
OlmoEarth v1.3 Base
& \textbf{97.2} & 95.7 \\
DINOv3 ViT-L/16
& \textbf{97.3} & 96.3 \\
\midrule
Mean
& 94.5 & \textbf{95.1} \\
\bottomrule
\end{tabular*}
\caption{
Scene classification on the EuroSAT spatial split:
test accuracy (\%).
Linear denotes a linear classifier trained on pooled
frozen-encoder features.
\textbf{Bold} indicates the best method per backbone, mean denotes the average over all six backbones.
}
\label{tab:eurosat-comparison}
\end{table}
\paragraph{Scene classification.}
Tab.~\ref{tab:eurosat-comparison} evaluates EuroSAT classification
with spatially separated training and test regions.
Our method achieves 95.1\% average accuracy across six frozen
backbones, exceeding the trained linear probe at 94.5\%.
It performs better with CROMA, Scale-MAE, and TerraMind,
while the linear probe leads with DOFA, OlmoEarth, and DINOv3.
These results extend the applicability of our approach from
dense prediction to image-level classification.

\subsection{Data-Scarce and Few-Shot Regimes}
\label{sec:data_scarce}


\newcommand{\sval}[2]{\ensuremath{#1\,\pm\,#2}}
\newcommand{\sbest}[2]{{\boldmath$#1\,\pm\,#2$}}

\begin{table*}[t]
\centering
\scriptsize
\setlength{\tabcolsep}{2.5pt}
\renewcommand{\arraystretch}{1.12}

\begin{tabular*}{\textwidth}{
    @{\extracolsep{\fill}}l
    cc @{\hspace{5pt}}|@{\hspace{5pt}}
    cc @{\hspace{5pt}}|@{\hspace{5pt}}
    cc @{\hspace{5pt}}|@{\hspace{5pt}}
    cc@{}
}
\toprule
& \multicolumn{2}{c}{HLS Burn Scars}
& \multicolumn{2}{c}{MADOS}
& \multicolumn{2}{c}{PASTIS}
& \multicolumn{2}{c}{xView2} \\
\cmidrule(lr){2-3}
\cmidrule(lr){4-5}
\cmidrule(lr){6-7}
\cmidrule(lr){8-9}
Backbone
& UPerNet & Ours
& UPerNet & Ours
& UPerNet & Ours
& UPerNet & Ours \\
\midrule

\multicolumn{9}{@{}l}{
    \textit{(a) 10\% training labels; xView2: 1\%}
} \\
\addlinespace[2pt]
CROMA optical
& \sval{81.0}{0.2} & \sbest{82.1}{0.3}
& \sval{33.8}{7.1} & \sbest{56.7}{5.3}
& \sval{18.7}{2.8} & \sbest{34.0}{1.6}
& \sbest{31.8}{2.1} & \sval{31.2}{2.1} \\
DOFA
& \sval{77.6}{1.5} & \sbest{81.5}{0.9}
& \sval{31.1}{6.0} & \sbest{58.8}{5.8}
& \sval{12.8}{0.8} & \sbest{28.7}{1.6}
& \sbest{40.7}{2.1} & \sval{34.2}{1.4} \\
Scale-MAE
& \sval{69.6}{1.3} & \sbest{81.5}{1.0}
& \sval{30.0}{2.4} & \sbest{54.4}{4.6}
& \sval{6.0}{0.4} & \sbest{28.0}{1.5}
& \sbest{40.8}{1.3} & \sval{31.8}{2.1} \\
TerraMind 1.0 Base
& \sval{77.9}{0.6} & \sbest{82.1}{1.2}
& \sval{30.5}{1.6} & \sbest{55.4}{4.7}
& \sval{17.8}{1.7} & \sbest{31.6}{1.4}
& \sbest{38.9}{3.1} & \sval{32.4}{2.1} \\
OlmoEarth v1.3 Base
& \sbest{84.2}{0.9} & \sval{84.0}{0.7}
& \sval{51.3}{4.5} & \sbest{59.3}{4.4}
& \sval{25.0}{3.2} & \sbest{39.2}{1.9}
& \sbest{32.4}{2.4} & \sval{31.1}{1.4} \\
DINOv3 ViT-L/16
& \sval{82.0}{0.9} & \sbest{83.2}{0.4}
& \sval{31.5}{1.3} & \sbest{61.1}{3.1}
& \sval{13.7}{1.2} & \sbest{29.3}{1.4}
& \sbest{45.1}{2.5} & \sval{44.3}{1.1} \\
\midrule
Mean
& 78.7 & \textbf{82.4}
& 34.7 & \textbf{57.6}
& 15.7 & \textbf{31.8}
& $\mathbf{38.3}$ & $34.2$ \\
\midrule

\multicolumn{9}{@{}l}{
    \textit{(b) 1\% training labels; xView2: 0.1\%}
} \\
\addlinespace[2pt]
CROMA optical
& \sval{64.9}{5.2} & \sbest{69.0}{11.4}
& \sval{16.0}{1.6} & \sbest{32.2}{4.2}
& \sval{8.0}{2.5} & \sbest{20.9}{1.0}
& \sval{25.6}{2.7} & \sbest{26.5}{1.2} \\
DOFA
& \sval{63.1}{3.4} & \sbest{68.8}{11.3}
& \sval{9.2}{1.5} & \sbest{32.5}{3.7}
& \sval{5.6}{0.6} & \sbest{17.9}{0.6}
& \sbest{31.0}{1.4} & \sval{28.5}{1.4} \\
Scale-MAE
& \sval{57.4}{0.7} & \sbest{69.4}{11.7}
& \sval{11.9}{2.6} & \sbest{31.9}{4.6}
& \sval{3.6}{0.1} & \sbest{18.0}{0.7}
& \sbest{31.5}{0.7} & \sval{26.2}{1.4} \\
TerraMind 1.0 Base
& \sval{68.1}{2.2} & \sbest{69.0}{11.0}
& \sval{11.3}{2.1} & \sbest{31.9}{4.1}
& \sval{7.1}{1.8} & \sbest{19.5}{0.9}
& \sbest{29.1}{0.8} & \sval{26.8}{1.6} \\
OlmoEarth v1.3 Base
& \sbest{78.7}{2.9} & \sval{72.6}{8.4}
& \sval{16.9}{2.0} & \sbest{33.3}{4.3}
& \sval{10.0}{3.4} & \sbest{27.8}{0.6}
& \sval{25.8}{2.5} & \sbest{26.1}{0.9} \\
DINOv3 ViT-L/16
& \sval{68.0}{1.9} & \sbest{69.4}{11.5}
& \sval{8.0}{0.2} & \sbest{31.0}{7.2}
& \sval{6.2}{1.1} & \sbest{18.0}{0.6}
& \sval{35.5}{3.5} & \sbest{37.1}{1.7} \\
\midrule
Mean
& 66.7 & \textbf{69.7}
& 12.2 & \textbf{32.1}
& 6.8 & \textbf{20.4}
& $\mathbf{29.8}$ & $28.5$ \\
\midrule

\addlinespace[6pt]
\multicolumn{9}{@{}l}{
    \textit{(c) Reference methods using DINOv3 ViT-L/16}
} \\
\addlinespace[2pt]
\midrule
& \multicolumn{2}{c}{HLS Burn Scars}
& \multicolumn{2}{c}{MADOS}
& \multicolumn{2}{c}{PASTIS}
& \multicolumn{2}{c}{xView2} \\
\cmidrule(lr){2-3}
\cmidrule(lr){4-5}
\cmidrule(lr){6-7}
\cmidrule(lr){8-9}
Method
& 10\% & 1\%
& 10\% & 1\%
& 10\% & 1\%
& 1\% & 0.1\% \\
\midrule
FROST \cite{park2026frost}
& \sval{63.0}{0.7} & \sval{50.8}{8.8}
& \sval{26.5}{7.1} & \sval{12.2}{1.3}
& \sval{12.1}{0.7} & \sval{7.6}{0.7}
& \sval{35.9}{0.8} & \sval{35.2}{1.2} \\
INSID3 \cite{cuttano2026insid3}
& \sval{49.4}{0.4} & \sval{38.3}{19.2}
& \sval{22.4}{3.5} & \sval{13.2}{1.1}
& \sval{11.0}{0.1} & \sval{7.4}{0.1}
& \sval{29.1}{0.7} & \sval{28.6}{0.5} \\
\bottomrule
\end{tabular*}

\caption{
Sparse-label evaluation: test mIoU (\%), reported as mean
$\pm$ sample standard deviation over three seeds.
Panels (a) and (b) compare UPerNet with our method;
panel (c) reports FROST and INSID3 with retrieved references.
\textbf{Bold} marks the higher mean within each completed
UPerNet/Ours pair and the better backbone-average result.
Mean averages the per-backbone means over the same backbones
for both methods.
}
\label{tab:sparse-backbone-comparison}
\end{table*}

Tab.~\ref{tab:sparse-backbone-comparison} shows that our method
achieves higher backbone-averaged performance than \textit{UPerNet} on
HLS, MADOS, and PASTIS at both label budgets, with particularly
large gains on MADOS and PASTIS.
These gains may reflect reduced risk of overfitting compared
with training high-capacity decoders on limited labelled data.
Our DINOv3 variant also consistently outperforms the established
training-free few-shot methods FROST~\cite{park2026frost}
and INSID3~\cite{cuttano2026insid3} across all evaluated datasets
and label budgets, often by a wide margin.
On xView2 in the few-shot regime (0.1\% labels), it sets the state-of-the-art among all evaluated backbones and methods (37.1\%), reversing
\textit{UPerNet}'s advantage in the full-label setting.

\subsection{Ablations and Analysis}
\begin{table}[t]
    \centering
    \begin{tabular}{@{}lrrr@{}}
        \toprule
        Configuration & DINOv3 & TerraMind & OlmoEarth \\
        \midrule
        \multicolumn{4}{@{}l}{\textit{Feature ablation}} \\
        Visual only       & 51.9          & 47.5          & 66.9          \\
        Visual + spectral & 68.0          & 63.6          & 68.8          \\
        Visual + pixel    & \textbf{79.8} & 69.3          & 73.3          \\
        Full features              & 79.3          & \textbf{71.6} & \textbf{75.8} \\
        \midrule
        \multicolumn{4}{@{}l}{\textit{Gaussianization ablation (full features)}} \\
        $K=0$             & 76.9          & 66.3          & 73.1          \\
        $K=1$             & \textbf{79.3} & \textbf{71.6} & \textbf{75.8} \\
        $K=2$             & 77.6          & 68.2          & 73.1          \\
        $K=3$             & 78.3          & 69.2          & 75.4          \\
        $K=5$             & 78.7          & 69.1          & \textbf{75.8} \\
        $K=10$            & 78.9          & 67.4          & 73.6          \\
        \bottomrule
    \end{tabular}
    \caption{Feature and Gaussianization ablations on MADOS
    with Gaussian-kernel token prediction.
    Top: spectral token statistics and the pixel branch are
    added independently to frozen visual embeddings; Full
    combines all three branches.
    Bottom: varying $K$ with the full feature set.
    Values are test mIoU (\%); the best score per backbone
    within each block is shown in bold.}
    \label{tab:mados_kernel_ablations}
\end{table}
We investigate how Gaussianization and complementary feature
branches improve prediction, and examine their effect on
positional bias in frozen representations.
Quantitative ablations use MADOS with Gaussian-kernel prediction
and three pretrained backbones.
Further ablations are provided in
Appendix~\ref{app:additional_results}.

\paragraph{Effect of Gaussianization.}
Fig.~\ref{fig:token_space_analysis} shows how the successive steps of
our transform reshape the frozen tokens toward an isotropic Gaussian:
the covariance spectrum flattens, the one-dimensional projections
align with a Gaussian, and the descriptor lengths concentrate. We
quantify this with the Epps--Pulley (EP)
statistic~\cite{epps1983normality}, which measures how far the tokens
are from a Gaussian (see Appendix~\ref{app:theoretical_details} for the
definition). From z-scoring to radial correction, it decreases from
$2.4\times10^{-2}$ to $5.5\times10^{-4}$, and mIoU improves from 67.6\%
to 75.7\%. The full transform is also the best setting for all three
backbones (Fig.~\ref{fig:mados_steps_ablation_appendix}, Appendix~\ref{sec:kernel-preprocessing}).

Tab.~\ref{tab:mados_kernel_ablations} varies the number of marginal
Gaussianization iterations $K$. A single iteration ($K=1$) achieves the
best or tied-best result for all three backbones. Further iterations
offer no consistent benefit, while all tested $K\geq1$ match or exceed
$K=0$. On other datasets, additional iterations occasionally help, but
only marginally, so we fix $K=1$ throughout. Consistent with the
empirical observations of Laparra et al.~\cite{laparra2011rbig}, we
find no practical evidence of PCA stalling.

The benefit of Gaussianization is not limited to learned features: for
the handcrafted spectral descriptors alone, it raises mIoU from 52.1\%
to 61.1\% (Tab.~\ref{tab:mados_kernel_ablations_app}, Appendix~\ref{app:spec_gauss}).
Appendix~\ref{sec:kernel-preprocessing} further compares our transform
with standard scaling in data-scarce and few-shot regimes.

\paragraph{Complementary feature branches.}
Adding either spectral statistics or the pixel branch improves
all three backbones over visual features alone, with larger
gains from the pixel branch, as shown in Tab.~\ref{tab:mados_kernel_ablations}.
Combining all branches performs best for TerraMind and
OlmoEarth.
For DINOv3, visual-plus-pixel features perform slightly better
than the full combination (79.8\% versus 79.3\%), indicating
that the benefit of spectral features depends on the backbone.

\begin{figure}[th]
    \centering
    \includegraphics[width=\columnwidth]{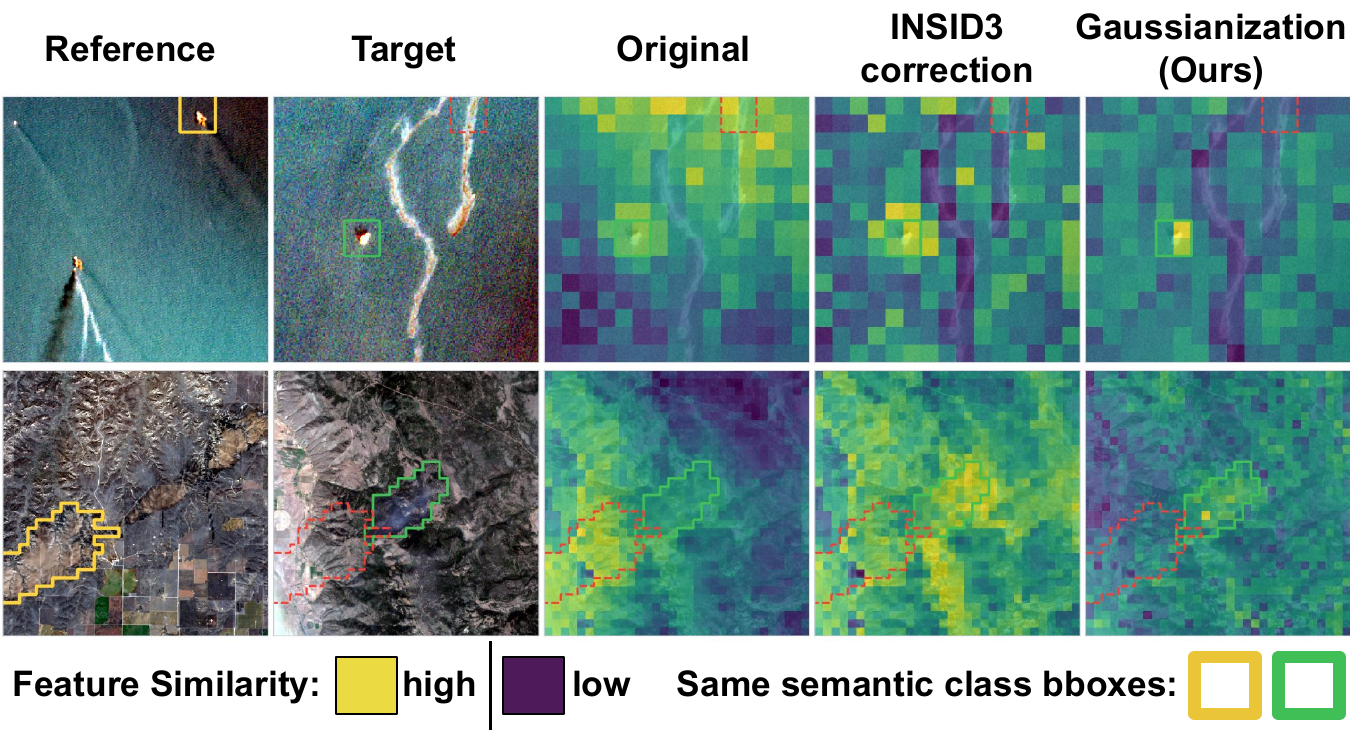}
\caption{Positional bias in DINOv3 cross-image similarity maps on MADOS (top) and HLSBurnScars (bottom), for original features, INSID3 debiasing~\cite{cuttano2026insid3}, and our Gaussianization. Dashed red outlines mark the reference region's coordinates in the target image. Ideally, similarity is high inside the green region and low inside the red one.}
    \label{fig:position_bias}
\end{figure}

\paragraph{Implicit positional debiasing.}
Foundation-model representations can retain spatial biases
when positional information becomes entangled with semantic
features, causing unrelated patches at similar image locations
to appear similar.
This poses a particular challenge for training-free few-shot
segmentation, where predictions rely directly on feature
similarities without gradient-based adaptation that could
compensate for these biases.
Established methods such as INSID3~\cite{cuttano2026insid3}
and FROST~\cite{park2026frost} therefore include explicit
positional debiasing steps.
Figure~\ref{fig:position_bias} illustrates this effect in DINOv3:
original features can exhibit high similarity at corresponding
spatial positions even for semantically unrelated regions.
Our Gaussianization pipeline attenuates these position-driven
responses and better highlights semantically corresponding
regions, without requiring a dedicated bias-correction module.
%
Appendix~\ref{app:position_bias} repeats the
comparison for the EO backbone TerraMind.

\subsection{Computational Cost}
Tab.~\ref{tab:compute} reports the end-to-end computational
cost on the MADOS dataset with the DINOv3 backbone, including
preparation, validation-based model selection, and test inference,
both on a GPU and on a CPU without any accelerator.
Trained-decoder baselines, which fit a decoder on top of the frozen
encoder, rely on gradient-based optimization that dominates their
preparation cost, although inference is efficient once training is
complete.
FROST~\cite{park2026frost} and INSID3~\cite{cuttano2026insid3} avoid
decoder training, but require reference-dependent matching and mask
construction for every query. In contrast, our method computes the
feature transformations and the reference bank once and amortizes them
over all subsequent predictions.
On a GPU, our method is about 3$\times$ faster than UPerNet and also
faster than the light decoder. The advantage persists in the CPU-only
setting, where our method finishes in 6.3 hours while UPerNet requires
an estimated 32 hours. It is even larger with an EO backbone: with
OlmoEarth, our method is almost 6$\times$ faster than UPerNet on a GPU (Tab.~\ref{tab:compute_mados_olmoearth}).
Combined with its strong performance in data-scarce settings, this
makes our method well suited for fast deployment of state-of-the-art
foundation models on modest hardware. 
Our peak memory is currently higher than that of the trained decoders,
which reflects an unoptimized implementation that keeps the full
reference bank in memory; we discuss this, together with further
results, in Appendix~\ref{app:compute}.


\begin{table}[t]
\centering
\small
\setlength{\tabcolsep}{4pt}
\begin{tabular}{@{}lrrrr@{}}
\toprule
 & \multicolumn{2}{c}{GPU} & \multicolumn{2}{c}{CPU-only} \\
\cmidrule(lr){2-3}\cmidrule(l){4-5}
Method & Time & VRAM & Time & RAM \\
       & (min) & (GB) & (h) & (GB) \\
\midrule
Ours (kNN)          & 12.0      & 10.2 & 6.3              & 17.0 \\
Light & 18.8      & 2.3  & 11.5$^\dagger$   & 3.1 \\
UPerNet      & 33.8      & 4.1  & 31.9$^\dagger$   & 4.3 \\
INSID3~\cite{cuttano2026insid3}        & 45.7      & 8.5  & 98.8$^\dagger$   & 10.7 \\
FROST~\cite{park2026frost}          & 1{,}009   & 24.4 & 1{,}340$^\dagger$ & 51.3 \\
\bottomrule
\end{tabular}
\caption{Computational cost on MADOS (1{,}433/642/728 train/val/test images) with a DINOv3 ViT-L/16 backbone. Time covers all steps: training (or reference-bank construction), parameter selection on the validation set, and inference on the test set. Memory is the peak usage on the respective device. GPU: one NVIDIA RTX PRO 6000 Blackwell MIG slice with 8 CPU cores; CPU: 8 cores of an AMD EPYC 7763 without an accelerator. $^\dagger$Extrapolated from short CPU runs.}
\label{tab:compute}
\end{table}

\section{Related Work}
\label{sec:related_work}

\paragraph{Foundation models for Earth observation.} Visual foundation models such as  DINOv3~\cite{simeoni2025dinov3} learn reusable dense representations through large scale self-supervised pretraining. EO-specific models adapt this paradigm to satellite imagery: CROMA~\cite{fuller2023croma} combines optical and radar observations, DOFA~\cite{xiong2024dofa} accommodates heterogeneous spectral inputs, and Prithvi~\cite{jakubik2023foundation} and SpectralGPT~\cite{hong2024spectralgpt} target multitemporal and spectral representations, respectively. PANGAEA~\cite{marsocci2026pangaea} documents substantial variation
in transfer performance across EO tasks and backbones. Turkoglu et al.~\cite{turkoglu2026reuse} further show that
fine-tuning can substantially alter the pretrained weight
structure of EO models, highlighting the distinction between
downstream accuracy and representation reuse. TerraMind~\cite{jakubik2025terramind} introduces generative cross-modal
modeling, while OlmoEarth~\cite{herzog2026olmoearth} develops
self-supervised spatiotemporal representations.
TESSERA~\cite{feng2026tessera} and
AlphaEarth Foundations~\cite{brown2025alphaearth} provide global
embedding maps derived from temporal and multisensor observations.

\paragraph{Kernel prediction and reference banks.}
Predicting a label as a similarity-weighted average of labeled examples dates back to Nadaraya--Watson regression~\cite{nadaraya1964estimating,watson1964smooth}. Weighted $k$NN voting is also a common evaluation protocol for self-supervised representations~\cite{wu2018unsupervised,caron2021emerging}. Hummingbird~\cite{balazevic2023hummingbird} extends reference-based prediction to dense scene understanding through retrieval from annotated feature memories. kNN-CLIP~\cite{gui2024knnclip} augments open-vocabulary segmentation
with a database of instance embeddings. INSID3~\cite{cuttano2026insid3} performs in-context segmentation with frozen DINOv3 features, while FROST~\cite{park2026frost} compares foreground and background density estimates for few-shot remote sensing
segmentation. We study joint multiclass EO prediction, based on labeled reference banks that combine visual features with explicit spectral and temporal descriptors.

\paragraph{Feature geometry and Gaussianization.} Isotropic Gaussian representations combine balanced covariance with distributional structure beyond second-order statistics. VICReg~\cite{bardes2022vicreg} regularizes feature variance and covariance, while LeJEPA~\cite{balestriero2025lejepa} motivates an isotropic Gaussian target through downstream-probe analysis, including bounds on kernel-regression bias under stated assumptions, and encourages this target during training with SIGReg. Classical Gaussianization provides a post-hoc approach to distribution matching. RBIG~\cite{laparra2011rbig} alternates marginal Gaussianization with rotations to address higher-order dependencies. Radial Gaussianization~\cite{lyu2009radial} calibrates vector lengths while preserving directions, yielding a joint Gaussian under spherical symmetry. These approaches inspired us to study the distributional calibration of visual representations. 
\section{Conclusion}\label{sec:conclusion}
We challenge the conventional approach to adapting Earth observation foundation models, which relies on fine-tuning or training high-capacity decoders. Our hypothesis is that pretrained features already contain much of the information needed for downstream tasks, but their geometry makes it difficult to access. We show that post-hoc Gaussianization makes these features suitable for simple distance-based prediction, without gradient-based optimization. The same framework also allows us to incorporate complementary information, including spectral features and raw pixel values. Across six benchmarks and multiple foundation models, we achieve competitive or superior performance, with particularly strong results when labels are scarce. This suggests that effective downstream adaptation need not involve additional training. Moreover, the ability to incorporate new labeled examples without retraining offers a natural starting point for agentic EO systems that accumulate and reuse knowledge through persistent reference banks.

\section*{LLM Usage Statement}
ChatGPT (OpenAI) was used as a writing aid for the manuscript to improve grammar. Claude Code (Anthropic) and Codex (OpenAI) were used as coding assistants during software development.

{
    \small
    \bibliographystyle{ieeenat_fullname}
    \bibliography{main}
}

\clearpage

\appendix
\appendix

\section{Additional Results}
\label{app:additional_results}
In this section, we present additional experimental results.

\subsection{Data-Scarce and Few-Shot Results}
Table~\ref{tab:sparse-backbone-comparison-light} extends the comparison
in Sec.~\ref{sec:data_scarce} with the \textit{Light} decoder and reports all three
methods per backbone.
Averaged across backbones, our method achieves the highest mIoU on HLS Burn Scars, MADOS, and PASTIS in both the data-scarce and the few-shot regime. Its margin over the stronger trained decoder is 3.7, 15.7, and 9.4 points at 10\% labels, and 3.0, 17.8, and 11.2 points at 1\%.

xView2 is the only dataset where our method is not the best on average. The gap to \textit{UPerNet} nevertheless shrinks as labels decrease, from 10.6 points with full labels (Tab.~\ref{tab:backbone-comparison-four-datasets}) to 4.1 points at 1\% and 1.3 points at 0.1\%.With DINOv3, our method is on par with \textit{UPerNet} at 1\% (44.3
vs.\ 45.1) and sets the state of the art among the evaluated methods at 0.1\% (37.1), ahead of every other backbone-decoder combination.
DINOv3 is also the strongest backbone on xView2 for both \textit{UPerNet} and our method, ahead of all EO encoders, plausibly because high-resolution imagery lies closer to its pretraining distribution.

\begin{table*}[t]
\centering
\footnotesize
\setlength{\tabcolsep}{2.5pt}
\renewcommand{\arraystretch}{1.10}


\newcommand{\SparseNumber}[1]{\makebox[2em][r]{#1}}
\newcommand{\SparseSpread}[1]{\hspace{0.23em}\ensuremath{\pm}\hspace{0.23em}\makebox[2em][l]{#1}}
\newcommand{\SparseScore}[2]{\SparseNumber{#1}\SparseSpread{#2}}
\newcommand{\SparseBest}[2]{\SparseScore{\textbf{#1}}{#2}}
\newcommand{\SparseAverage}[1]{\SparseNumber{#1}\phantom{\SparseSpread{00.0}}}
\newcommand{\SparseHeader}[5]{%
  \textbf{#1} & \multicolumn{3}{c}{\textbf{#2\% labels}~(\(n=#3\))}
              & \multicolumn{3}{c@{}}{\textbf{#4\% labels}~(\(n=#5\))} \\
  \cmidrule(lr){2-4}\cmidrule(l){5-7}
  \textit{Backbone} & Light & UPerNet & Ours & Light & UPerNet & Ours \\
  \addlinespace[1pt]
}

\begin{tabular*}{\textwidth}{@{\extracolsep{\fill}}lccc@{\hspace{1.5em}}ccc@{}}
\toprule
\SparseHeader{HLS Burn Scars}{10}{48}{1}{4}
CROMA optical & \SparseScore{78.5}{0.6} & \SparseScore{81.0}{0.2} & \SparseBest{82.1}{0.3}
  & \SparseScore{63.9}{8.2} & \SparseScore{64.9}{5.2} & \SparseBest{69.0}{11.4} \\
DOFA & \SparseScore{71.7}{1.6} & \SparseScore{77.6}{1.5} & \SparseBest{81.5}{0.9}
  & \SparseScore{48.6}{5.5} & \SparseScore{63.1}{3.4} & \SparseBest{68.8}{11.3} \\
Scale-MAE & \SparseScore{65.1}{1.5} & \SparseScore{69.6}{1.3} & \SparseBest{81.5}{1.0}
  & \SparseScore{48.6}{4.4} & \SparseScore{57.4}{0.7} & \SparseBest{69.4}{11.7} \\
TerraMind 1.0 Base & \SparseScore{76.9}{0.9} & \SparseScore{77.9}{0.5} & \SparseBest{82.1}{1.2}
  & \SparseScore{58.1}{7.2} & \SparseScore{68.1}{2.2} & \SparseBest{69.0}{11.0} \\
OlmoEarth v1.3 Base & \SparseScore{83.6}{0.1} & \SparseBest{84.2}{0.9} & \SparseScore{84.0}{0.7}
  & \SparseScore{66.6}{12.3} & \SparseBest{78.7}{2.9} & \SparseScore{72.6}{8.4} \\
DINOv3 ViT-L/16 & \SparseScore{79.5}{1.3} & \SparseScore{82.0}{0.9} & \SparseBest{83.2}{0.4}
  & \SparseScore{51.1}{9.0} & \SparseScore{68.0}{1.9} & \SparseBest{69.4}{11.5} \\

\addlinespace[2pt]
\textbf{Mean across backbones} & \SparseAverage{75.9} & \SparseAverage{78.7} & \SparseAverage{\textbf{82.4}}
  & \SparseAverage{56.2} & \SparseAverage{66.7} & \SparseAverage{\textbf{69.7}} \\
\midrule

\SparseHeader{MADOS}{10}{143}{1}{14}
CROMA optical & \SparseScore{40.0}{1.8} & \SparseScore{33.8}{7.1} & \SparseBest{56.7}{5.3}
  & \SparseScore{19.9}{3.9} & \SparseScore{16.0}{1.6} & \SparseBest{32.2}{4.2} \\
DOFA & \SparseScore{36.7}{2.5} & \SparseScore{31.1}{6.0} & \SparseBest{58.8}{5.8}
  & \SparseScore{10.2}{2.4} & \SparseScore{9.2}{1.5} & \SparseBest{32.5}{3.7} \\
Scale-MAE & \SparseScore{33.5}{1.6} & \SparseScore{30.0}{2.4} & \SparseBest{54.4}{4.6}
  & \SparseScore{12.7}{4.3} & \SparseScore{11.9}{2.6} & \SparseBest{31.9}{4.6} \\
TerraMind 1.0 Base & \SparseScore{41.4}{4.0} & \SparseScore{30.5}{1.6} & \SparseBest{55.4}{4.7}
  & \SparseScore{13.6}{1.5} & \SparseScore{11.3}{2.1} & \SparseBest{31.9}{4.1} \\
OlmoEarth v1.3 Base & \SparseScore{58.0}{4.0} & \SparseScore{51.3}{4.5} & \SparseBest{59.3}{4.4}
  & \SparseScore{20.4}{4.6} & \SparseScore{16.9}{2.0} & \SparseBest{33.3}{4.3} \\
DINOv3 ViT-L/16 & \SparseScore{41.8}{2.4} & \SparseScore{31.5}{1.3} & \SparseBest{61.1}{3.1}
  & \SparseScore{9.1}{1.9} & \SparseScore{7.9}{0.2} & \SparseBest{31.0}{7.2} \\

\addlinespace[2pt]
\textbf{Mean across backbones} & \SparseAverage{41.9} & \SparseAverage{34.7} & \SparseAverage{\textbf{57.6}}
  & \SparseAverage{14.3} & \SparseAverage{12.2} & \SparseAverage{\textbf{32.1}} \\
\midrule

\SparseHeader{PASTIS}{10}{145}{1}{14}
CROMA optical & \SparseScore{26.0}{1.0} & \SparseScore{18.7}{2.8} & \SparseBest{34.0}{1.6}
  & \SparseScore{11.5}{1.9} & \SparseScore{8.0}{2.5} & \SparseBest{20.9}{1.0} \\
DOFA & \SparseScore{18.1}{0.1} & \SparseScore{12.8}{0.8} & \SparseBest{28.7}{1.6}
  & \SparseScore{7.0}{0.9} & \SparseScore{5.6}{0.6} & \SparseBest{17.9}{0.6} \\
Scale-MAE & \SparseScore{13.7}{0.4} & \SparseScore{6.0}{0.4} & \SparseBest{28.0}{1.5}
  & \SparseScore{6.2}{1.0} & \SparseScore{3.6}{0.1} & \SparseBest{18.0}{0.7} \\
TerraMind 1.0 Base & \SparseScore{25.7}{0.9} & \SparseScore{17.8}{1.7} & \SparseBest{31.6}{1.4}
  & \SparseScore{8.7}{1.8} & \SparseScore{7.1}{1.8} & \SparseBest{19.5}{0.9} \\
OlmoEarth v1.3 Base & \SparseScore{31.4}{1.8} & \SparseScore{25.0}{3.2} & \SparseBest{39.2}{1.9}
  & \SparseScore{14.2}{2.2} & \SparseScore{10.0}{3.4} & \SparseBest{27.8}{0.6} \\
DINOv3 ViT-L/16 & \SparseScore{19.5}{1.0} & \SparseScore{13.7}{1.2} & \SparseBest{29.3}{1.4}
  & \SparseScore{7.6}{1.3} & \SparseScore{6.2}{1.1} & \SparseBest{18.0}{0.6} \\

\addlinespace[2pt]
\textbf{Mean across backbones} & \SparseAverage{22.4} & \SparseAverage{15.7} & \SparseAverage{\textbf{31.8}}
  & \SparseAverage{9.2} & \SparseAverage{6.8} & \SparseAverage{\textbf{20.4}} \\
\midrule

\SparseHeader{xView2}{1}{133}{0.1}{13}
CROMA optical & \SparseScore{24.8}{1.0} & \SparseBest{31.8}{2.1} & \SparseScore{31.2}{2.1}
  & \SparseScore{21.7}{1.5} & \SparseScore{25.6}{2.7} & \SparseBest{26.5}{1.2} \\
DOFA & \SparseScore{27.4}{1.0} & \SparseBest{40.7}{2.1} & \SparseScore{34.2}{1.4}
  & \SparseScore{21.1}{1.3} & \SparseBest{31.0}{1.3} & \SparseScore{28.5}{1.4} \\
Scale-MAE & \SparseScore{35.0}{0.8} & \SparseBest{40.8}{1.3} & \SparseScore{31.8}{2.1}
  & \SparseScore{24.3}{0.2} & \SparseBest{31.5}{0.7} & \SparseScore{26.2}{1.4} \\
TerraMind 1.0 Base & \SparseScore{27.6}{1.8} & \SparseBest{38.9}{3.1} & \SparseScore{32.4}{2.1}
  & \SparseScore{21.6}{0.2} & \SparseBest{29.1}{0.8} & \SparseScore{26.8}{1.6} \\
OlmoEarth v1.3 Base & \SparseScore{24.2}{0.7} & \SparseBest{32.4}{2.4} & \SparseScore{31.1}{1.4}
  & \SparseScore{21.6}{0.8} & \SparseScore{25.8}{2.5} & \SparseBest{26.1}{0.9} \\
DINOv3 ViT-L/16 & \SparseScore{29.2}{1.5} & \SparseBest{45.1}{2.5} & \SparseScore{44.3}{1.1}
  & \SparseScore{20.0}{1.4} & \SparseScore{35.5}{3.5} & \SparseBest{37.1}{1.7} \\

\addlinespace[2pt]
\textbf{Mean across backbones} & \SparseAverage{28.0} & \SparseAverage{\textbf{38.3}} & \SparseAverage{34.2}
  & \SparseAverage{21.7} & \SparseAverage{\textbf{29.8}} & \SparseAverage{28.5} \\
\bottomrule
\end{tabular*}

\caption{\textbf{Data-scarce and few-shot performance} for segmentation per backbone and
decoder. Scores are test mIoU (\%), mean $\pm$ standard deviation over
three seeds. \textbf{Bold} marks the best method per backbone and label budget.}
\label{tab:sparse-backbone-comparison-light}
\end{table*}

\subsection{Spectral Descriptors in Supervised Decoders}
Spectral descriptors clearly help our method (Tab.~\ref{tab:mados_kernel_ablations_app}). A natural question is whether trained decoders would benefit in the same way if they received the same input. We test this on MADOS and PASTIS by concatenating the per-patch spectral descriptors to the frozen backbone features before training the \textit{Light} and \textit{UPerNet} decoders.

Tab.~\ref{tab:spectral-descriptor-ablation} shows that the trained
decoders do not benefit from this additional input. Test mIoU drops in eight of the twelve comparisons, and by 0.8 and 0.7 points on average on MADOS and PASTIS. The effect is also inconsistent: TerraMind with \textit{UPerNet} loses 6.5 points on MADOS but gains 3.9 on PASTIS. This suggests that the gains in our method do not simply come from extra input information that the trained decoders lack.

\begin{table*}[t]
\centering
\small
\begin{tabular}{@{}llrrrrrr@{}}
\toprule
& & \multicolumn{3}{c}{MADOS} & \multicolumn{3}{c}{PASTIS} \\
\cmidrule(lr){3-5}\cmidrule(lr){6-8}
Backbone & Decoder & Backbone-only & + Spectral & $\Delta$ & Backbone-only & + Spectral & $\Delta$ \\
\midrule
DINOv3 ViT-L/16 & Light & \textbf{73.1} & 72.4 & $ -0.7 $ & 30.2 & \textbf{30.3} & $ +0.1 $ \\
 & UPerNet & \textbf{73.8} & 73.5 & $ -0.3 $ & \textbf{28.6} & 24.6 & $ -4.0 $ \\
\addlinespace
TerraMind Base & Light & 60.0 & \textbf{62.2} & $ +2.2 $ & \textbf{37.0} & 36.5 & $ -0.5 $ \\
 & UPerNet & \textbf{72.0} & 65.5 & $ -6.5 $ & 29.0 & \textbf{32.9} & $ +3.9 $ \\
\addlinespace
OlmoEarth v1.3 Base & Light & \textbf{74.5} & 74.3 & $ -0.2 $ & \textbf{41.4} & 40.4 & $ -1.0 $ \\
 & UPerNet & 79.5 & \textbf{80.3} & $ +0.8 $ & \textbf{39.0} & 36.4 & $ -2.6 $ \\
\midrule
\multicolumn{2}{@{}l}{Mean}
 & \textbf{72.2} & 71.4 & $-0.8$
 & \textbf{34.2} & 33.5 & $-0.7$ \\
 \bottomrule
\end{tabular}
\caption{\textbf{Effect of adding spectral descriptors to the trained
decoder baselines.} Scores are test mIoU (\%) without and with the
descriptors concatenated to the frozen backbone features. $\Delta$ is
the change in percentage points, computed before rounding.
\textbf{Bold} marks the higher score in each pair.}
\label{tab:spectral-descriptor-ablation}
\end{table*}

\subsection{Ablations}
\begin{table}[t]
    \centering
    \small
    \setlength{\tabcolsep}{4pt}
    \renewcommand{\arraystretch}{1.08}
    \resizebox{\columnwidth}{!}{%

    \begin{tabular}{@{}lrrr@{}}
        \toprule
        & \multicolumn{3}{c@{}}{Test mIoU (\%)} \\
        \cmidrule(l){2-4}
        Configuration & DINOv3 & TerraMind & OlmoEarth \\
        \midrule
        \multicolumn{4}{@{}l}{\textit{Feature ablation with SLIC} ($K=1$)} \\
        \addlinespace[2pt]
        Visual-only (wo/ SLIC)   & 50.1 & 47.0 & 67.1 \\
        Visual-only     & 51.9          & 47.5          & 66.9 \\
        Visual + spectral & 68.0          & 63.6          & 68.8 \\
        Visual + pixel    & \textbf{79.8} & 69.3          & 73.3 \\
        Full features     & 79.3          & \textbf{71.6} & \textbf{75.8} \\
        \midrule
        \addlinespace[6pt]
        \multicolumn{4}{@{}l}{\textit{Feature ablation without SLIC}} \\
        \addlinespace[2pt]
        
        Spectral-only ($K=0$) & \multicolumn{3}{c@{}}{52.1} \\
        Spectral-only ($K=1$) & \multicolumn{3}{c@{}}{61.1} \\
        Spectral-only ($K=1$ w/ SLIC)    & 
        \multicolumn{3}{c@{}}{61.4} \\
        \midrule
        \addlinespace[6pt]
        \multicolumn{4}{@{}l}{\textit{Gaussianization ablation} (full features)} \\
        \addlinespace[2pt]
        $K=0$  & 76.9          & 66.3          & 73.1 \\
        $K=1$  & \textbf{79.3} & \textbf{71.6} & \textbf{75.8} \\
        $K=2$  & 77.6          & 68.2          & 73.1 \\
        $K=3$  & 78.3          & 69.2          & 75.4 \\
        $K=5$  & 78.7          & 69.1          & \textbf{75.8} \\
        $K=10$ & 78.9          & 67.4          & 73.6 \\
        \bottomrule
    \end{tabular}
    }
   \caption{\textbf{Feature and Gaussianization ablations} on MADOS with
    Gaussian-kernel prediction. \emph{Full features} combines the visual,
    spectral, and pixel branches. Spectral-only scores need no encoder
    and span all three columns. \textbf{Bold} marks the best score per backbone
    in each block.}
    \label{tab:mados_kernel_ablations_app}
\end{table}

Tab.~\ref{tab:mados_kernel_ablations_app} extends the MADOS ablations
of Tab.~\ref{tab:mados_kernel_ablations}.

\paragraph{Feature branches.}
Using visual tokens alone yields 47.5--66.9\% mIoU. Adding spectral
token statistics improves this by 16.1 points for DINOv3 and TerraMind,
and by 1.9 points for OlmoEarth, whose visual-only result is already
the strongest. The pixel branch contributes more, with gains of
6.4--27.9 points, because it restores the spatial detail that patch
tokens lack. Note that the pixel branch is not independent of the
visual branch: it matches pixels only within reference patches
retrieved by visual descriptors. We therefore do not report a
spectral + pixel configuration. Combining all three branches is best
for TerraMind (71.6\%) and OlmoEarth (75.8\%), while DINOv3 peaks with
visual and pixel features (79.8\%). SLIC refinement has only a small
effect on MADOS, changing visual-only mIoU by $-0.2$ to $+1.8$ points.

\paragraph{Gaussianization of spectral descriptors.}
\label{app:spec_gauss}
The transform is not specific to learned features. With PCA whitening
only, the handcrafted spectral descriptors reach 52.1\% mIoU. Adding
the marginal and radial steps raises this to 61.1\%, a gain of 9.0
points, and to 61.4\% with SLIC. After Gaussianization, the spectral
descriptors alone therefore exceed the visual-only results of DINOv3
and TerraMind without any pretrained encoder.

\paragraph{Gaussianization rounds.}
A single round is enough. With radial correction kept at every $K$,
going from $K=0$ to $K=1$ improves mIoU by 2.4, 5.3, and 2.7 points for
DINOv3, TerraMind, and OlmoEarth, and $K=1$ is the best or tied-best
setting for every backbone. Further rounds bring no additional gain, so
we fix $K=1$ in all experiments. More rounds can help in individual cases, e.g., on xView2, but the gain is usually marginal.

\subsection{Effect of Gaussianization Stages}

\label{sec:kernel-preprocessing}
\begin{figure}[th]
    \centering
    \includegraphics[width=\columnwidth]
    {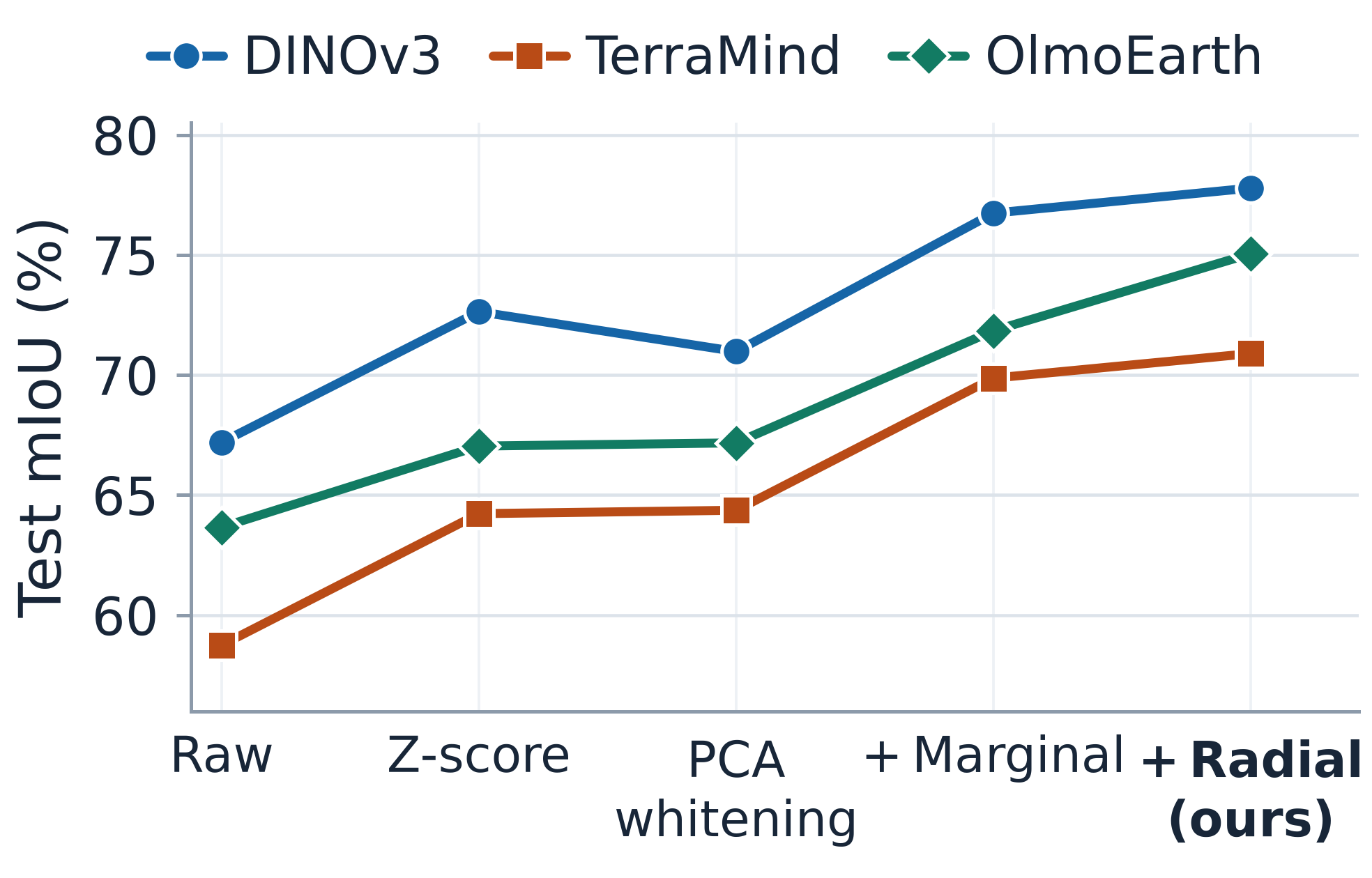}
\caption{\textbf{Effect of each Gaussianization stage} on MADOS (test mIoU,
\%). Only the transform applied to the token features changes. Stages
are added one at a time from left to right: raw features, z-scoring,
PCA whitening, marginal Gaussianization, and radial correction, which
gives our full transform.}
    \label{fig:mados_steps_ablation_appendix}
\end{figure}

\begin{table*}[t]
\centering
\small
\setlength{\tabcolsep}{4pt}
\renewcommand{\arraystretch}{1.10}
\begin{tabular*}{\textwidth}{@{\extracolsep{\fill}}lllrrrrrr@{}}
\toprule
\textbf{Dataset} & \textbf{Labels} & \textbf{Backbone}
& \multicolumn{2}{c}{\textbf{Raw}}
& \multicolumn{2}{c}{\textbf{Z-score}}
& \multicolumn{2}{c}{\textbf{Ours}} \\
\cmidrule(lr){4-5}\cmidrule(lr){6-7}\cmidrule(lr){8-9}
& & & EP & mIoU \(\uparrow\)
& EP & mIoU \(\uparrow\)
& EP & mIoU \(\uparrow\) \\
\midrule
MADOS & 1\% & DINOv3 ViT-L/16 & 375.5 & \(28.7 \pm 2.8\) & 20.1 & \(29.2 \pm 1.8\) & 12.7 & {\boldmath \(32.3 \pm 5.2\)} \\
 &  & TerraMind Base & 568.8 & \(28.5 \pm 2.0\) & 59.9 & \(27.6 \pm 0.3\) & 31.8 & {\boldmath \(31.2 \pm 4.9\)} \\
 &  & OlmoEarth v1.3 Base & 519.5 & \(31.2 \pm 1.2\) & 44.0 & \(30.6 \pm 2.8\) & 33.4 & {\boldmath \(33.0 \pm 6.4\)} \\
\addlinespace[1pt]
 &  & \textbf{Mean} & 487.9 & \(29.5\) & 41.3 & \(29.1\) & 26.0 & {\boldmath \(32.2\)} \\
\addlinespace[3pt]
 & 10\% & DINOv3 ViT-L/16 & 375.5 & \(55.4 \pm 4.2\) & 6.6 & \(55.7 \pm 2.5\) & 2.3 & {\boldmath \(61.5 \pm 4.4\)} \\
 &  & TerraMind Base & 568.0 & \(46.3 \pm 4.6\) & 61.4 & \(47.4 \pm 4.9\) & 6.1 & {\boldmath \(54.4 \pm 4.3\)} \\
 &  & OlmoEarth v1.3 Base & 519.5 & \(58.4 \pm 6.5\) & 31.3 & \(58.1 \pm 5.6\) & 6.3 & {\boldmath \(60.1 \pm 7.6\)} \\
\addlinespace[1pt]
 &  & \textbf{Mean} & 487.7 & \(53.4\) & 33.1 & \(53.7\) & 4.9 & {\boldmath \(58.7\)} \\
\midrule
PASTIS & 1\% & DINOv3 ViT-L/16 & 368.3 & \(9.7 \pm 6.4\) & 7.7 & \(14.7 \pm 1.0\) & 6.0 & {\boldmath \(15.5 \pm 0.4\)} \\
 &  & TerraMind Base & 377.0 & \(10.9 \pm 7.5\) & 6.6 & \(16.1 \pm 0.9\) & 3.4 & {\boldmath \(16.9 \pm 0.4\)} \\
 &  & OlmoEarth v1.3 Base & 385.7 & \(16.1 \pm 12.0\) & 5.5 & \(20.4 \pm 1.4\) & 4.8 & {\boldmath \(22.2 \pm 0.6\)} \\
\addlinespace[1pt]
 &  & \textbf{Mean} & 377.0 & \(12.2\) & 6.6 & \(17.1\) & 4.7 & {\boldmath \(18.2\)} \\
\addlinespace[3pt]
 & 10\% & DINOv3 ViT-L/16 & 368.3 & \(2.3 \pm 0.0\)\textsuperscript{\(\ddagger\)} & 1.4 & \(23.1 \pm 0.8\) & 1.2 & {\boldmath \(26.2 \pm 1.3\)} \\
 &  & TerraMind Base & 377.0 & \(2.3 \pm 0.0\)\textsuperscript{\(\ddagger\)} & 2.4 & \(25.9 \pm 0.9\) & 0.6 & {\boldmath \(28.6 \pm 1.7\)} \\
 &  & OlmoEarth v1.3 Base & 385.7 & \(2.3 \pm 0.0\)\textsuperscript{\(\ddagger\)} & 1.1 & \(32.6 \pm 1.2\) & 0.6 & {\boldmath \(34.6 \pm 1.5\)} \\
\addlinespace[1pt]
 &  & \textbf{Mean} & 377.0 & \(2.3\)\textsuperscript{\(\ddagger\)} & 1.6 & \(27.2\) & 0.8 & {\boldmath \(29.8\)} \\
\bottomrule
\end{tabular*}
\caption{\textbf{Gaussianization versus raw and z-scored token
features} in data-scarce and few-shot regimes. EP is the Epps--Pulley
score on held-out tokens (\(\times 10^{-3}\); lower is more Gaussian).
mIoU (\%) is the test mean \(\pm\) standard deviation over three label
subsets, without SLIC. \textbf{Bold} marks the best mIoU for each backbone and label budget. (\(\ddagger\)) At 10\% labels, raw PASTIS
features made the kernel bandwidth calibration numerically unstable.}
\label{tab:sparse-kernel-preprocessing}
\end{table*}

Gaussian-kernel prediction weights labeled references by their squared
Euclidean distance to the query~\cite{nadaraya1964estimating}, so
coordinates with large scales can dominate. Feature-wise scaling,
typically z-scoring, is the standard
remedy~\cite{hsu2025practical,sklearnStandardScaler}. We therefore test
whether our full transform is needed beyond z-scoring, and how much
each stage contributes. To isolate the transform, all runs in this
section omit SLIC refinement.

\paragraph{Individual stages.}
Fig.~\ref{fig:mados_steps_ablation_appendix} adds the stages one at a
time on MADOS with full labels. Z-scoring clearly improves over raw
features, and the complete transform is best for all three backbones.
PCA whitening on its own, in contrast, changes little and even lowers
the score slightly for DINOv3. This does not make it dispensable.
Whitening decorrelates the coordinates, already moves the features
closer to a Gaussian, and defines the axes along which marginal
Gaussianization acts. Its benefit becomes visible once the marginal
step is applied on top, which gives the largest gain of all stages for
every backbone, followed by a further gain from radial
correction. The stages are therefore coupled, and we always apply them
together.

\paragraph{Data-scarce regimes.}
Tab.~\ref{tab:sparse-kernel-preprocessing} compares raw, z-scored, and
fully transformed features on MADOS and PASTIS at 1\% and 10\% labels,
with three backbones and three label subsets per budget.
Our transform gives the highest mIoU in every setting, and
also the lowest Epps--Pulley
score~\cite{epps1983normality,balestriero2025lejepa}, measured on
held-out validation tokens (lower is more Gaussian). Compared with
z-scoring, mean mIoU increases by 3.1 and 5.0 points on MADOS at 1\%
and 10\% labels, and by 1.1 and 2.6 points on PASTIS, and the Epps--Pulley score is lower in every individual run, across
all datasets, label budgets, backbones, and label subsets.

\subsection{Choice of Predictor}
\label{sec:retrieval-predictor-choice}

Our method can predict with cosine \(k\)NN or with the Gaussian kernel.
At first sight the two are different: \(k\)NN makes a hard choice of
which references vote, while the kernel weights all references softly
by their distance to the query.
In theory, however, the two are equivalent for unit-normalized
descriptors in our Gaussianized space. As shown in
Appendix~\ref{app:theoretical_details}, the Gaussian kernel then
reduces to an exponential weight on cosine similarity with temperature
\(T=h^2\). Both predictors therefore rank the references in the same
order and, when they vote over the same references with these weights,
give the same prediction.
In practice, the two differ. The kernel is applied without unit
normalization, so it also depends on the descriptor norms, and the
number of neighbors, the weighting, and the bandwidth are tuned
separately for each predictor. Whether these differences matter for
accuracy is an empirical question.

Tab.~\ref{tab:probe-choice} shows that they matter little. For each
dataset and backbone, we tune both predictors independently on the
validation split (Appendix~\ref{app:selection}). With full labels and
averaged over backbones, they differ by at most 0.9 points on every
segmentation dataset and reach the same mean mIoU of 64.3\%. Selecting
the better one per dataset and backbone, as done for the main results,
gives 64.4\%.

\subsection{Computational Cost}
\label{app:compute}
\begin{table}[t]
\centering
\small
\setlength{\tabcolsep}{4pt}
\begin{tabular}{@{}lrrrr@{}}
\toprule
 & \multicolumn{2}{c}{GPU} & \multicolumn{2}{c}{CPU-only} \\
\cmidrule(lr){2-3}\cmidrule(l){4-5}
Method & Time & VRAM & Time & RAM \\
       & (min) & (GB) & (h) & (GB) \\
\midrule
Ours (kNN)          & 10.3      & 10.2 & 4.8              & 16.4 \\
Light & 17.3      & 1.0  & 14.1$^\dagger$   & 2.0 \\
UPerNet      & 60.1      & 7.2  & 94.4$^\dagger$   & 6.9 \\
INSID3$^\ddagger$~\cite{cuttano2026insid3}        & 45.7      & 8.5  & 98.8$^\dagger$   & 10.7 \\
FROST$^\ddagger$~\cite{park2026frost}          & 1{,}009   & 24.4 & 1{,}340$^\dagger$ & 51.3 \\
\bottomrule
\end{tabular}
\caption{\textbf{Computational cost} on MADOS (1{,}433/642/728 train/val/test images) with an OlmoEarth v1.3 Base backbone. Time covers all steps: training (or reference-bank construction), parameter selection on the validation set, and inference on the test set. Memory is the peak usage on the respective device. GPU: one NVIDIA RTX PRO 6000 Blackwell MIG slice with 8 CPU cores; CPU: 8 cores of an AMD EPYC 7763 without an accelerator. $^\dagger$Extrapolated from short CPU runs. $^\ddagger$With their own DINOv3 ViT-L/16 backbone, as in Tab.~\ref{tab:compute}.}
\label{tab:compute_mados_olmoearth}
\end{table}

Tab.~\ref{tab:compute_mados_olmoearth} repeats the MADOS cost comparison of Sec.~\ref{sec:exp} with OlmoEarth v1.3 Base, a recent EO foundation model. The efficiency advantage grows with this backbone.
Our method completes reference-bank construction, validation-based selection, and test inference in 10.3\,min on a GPU, almost $6\times$ faster than UPerNet (60.1\,min), compared with about $3\times$ for
DINOv3. Without an accelerator, the full pipeline takes 4.8\,h on eight CPU cores, against an estimated 94.4\,h for UPerNet. Adapting a frozen backbone to a new task is therefore feasible within minutes on a GPU and within hours on a CPU alone.

Peak memory is higher than for the trained decoders (10.2\,GB vs.\ 7.2\,GB for UPerNet and 1.0\,GB for Light on GPU). This reflects our unoptimized implementation, not a requirement of the method. The encoder only runs forward passes, and the peak arises from holding the full reference bank in memory for batched exhaustive search.
Approximate nearest-neighbor indexing and product  quantization~\cite{johnson2021billion,jegou2011product} could reduce both search time and memory, which we leave to future work.

\subsection{Implicit Positional Debiasing}
\label{app:position_bias}
\begin{figure}[th]
    \centering
    \includegraphics[width=\columnwidth]{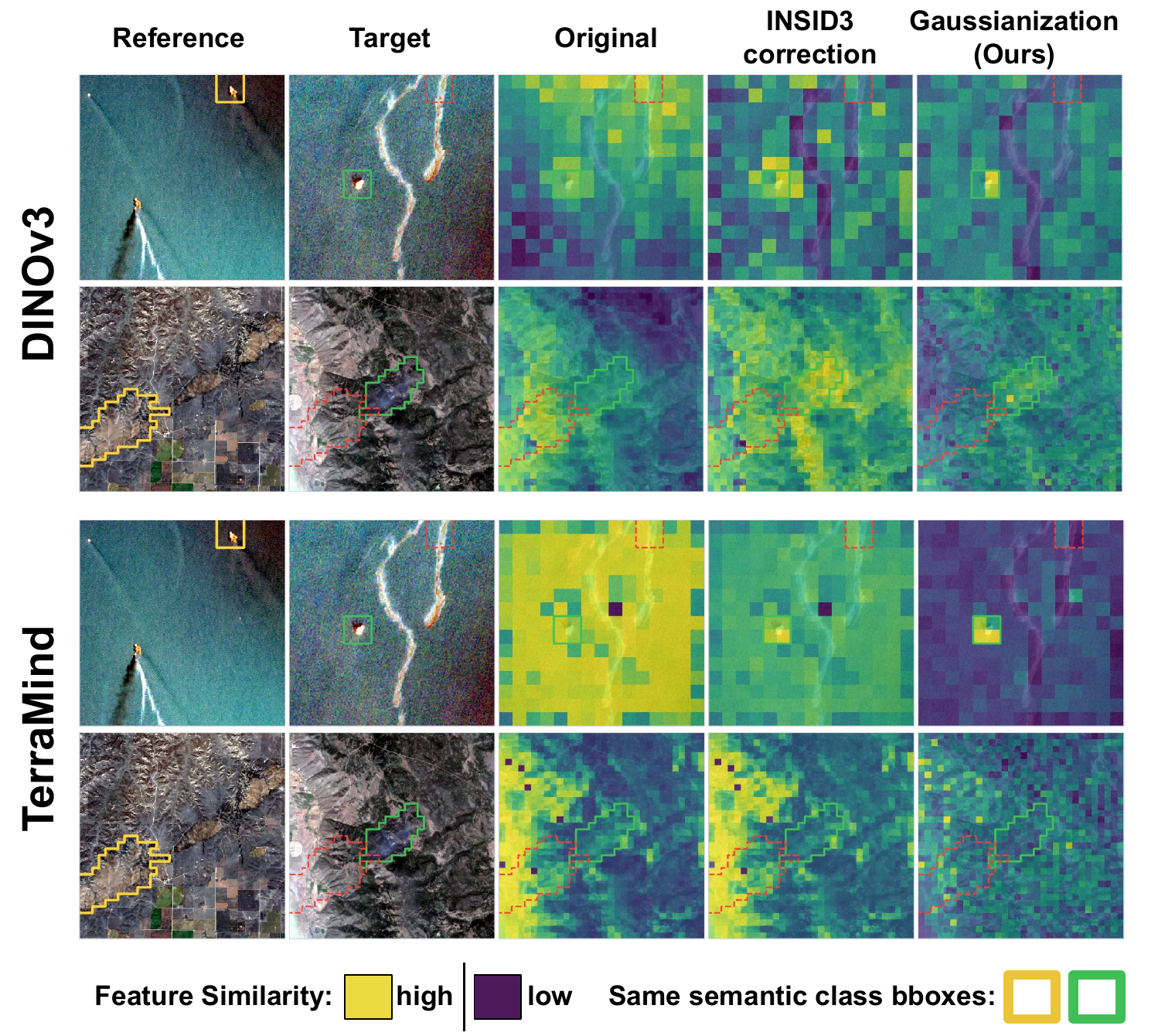}
\caption{Positional bias in cross-image similarity maps for DINOv3 and
TerraMind. For each backbone, the top row shows MADOS and the bottom
row HLS Burn Scars. Each map shows the similarity between the reference
region (yellow) and all target locations, for original features, INSID3
debiasing~\cite{cuttano2026insid3}, and our Gaussianization. The green
box marks the same class in the target, and the dashed red outline
marks where the reference region lies in image coordinates. Ideally,
similarity is high in the green box.}
    \label{fig:position_bias_full}
\end{figure}

Sec.~3.3 showed that DINOv3 features can assign high similarity to semantically unrelated regions that share image coordinates with the reference. 
Fig.~\ref{fig:position_bias_full} repeats this comparison for TerraMind and shows that the bias is not exclusive to DINOv3 features.
On HLS Burn Scars, the original TerraMind features respond most strongly along the left image border, which overlaps the reference coordinates and does not follow the burned area. The INSID3~\cite{cuttano2026insid3} correction leaves this pattern almost unchanged. Our Gaussianization removes it and yields a spatially more homogeneous similarity map.
On MADOS, TerraMind shows no clear positional pattern, but similarity is high across nearly the entire scene, so the region of the reference class does not stand out. Both INSID3~\cite{cuttano2026insid3} and Gaussianization restore this
contrast, and Gaussianization gives the sharper localization, with low similarity over the surrounding water. As for DINOv3, these effects are obtained without a dedicated bias-correction module.

\subsection{Superpixel Refinement}
\begin{figure*}[t]
    \centering
    \includegraphics[width=2\columnwidth]{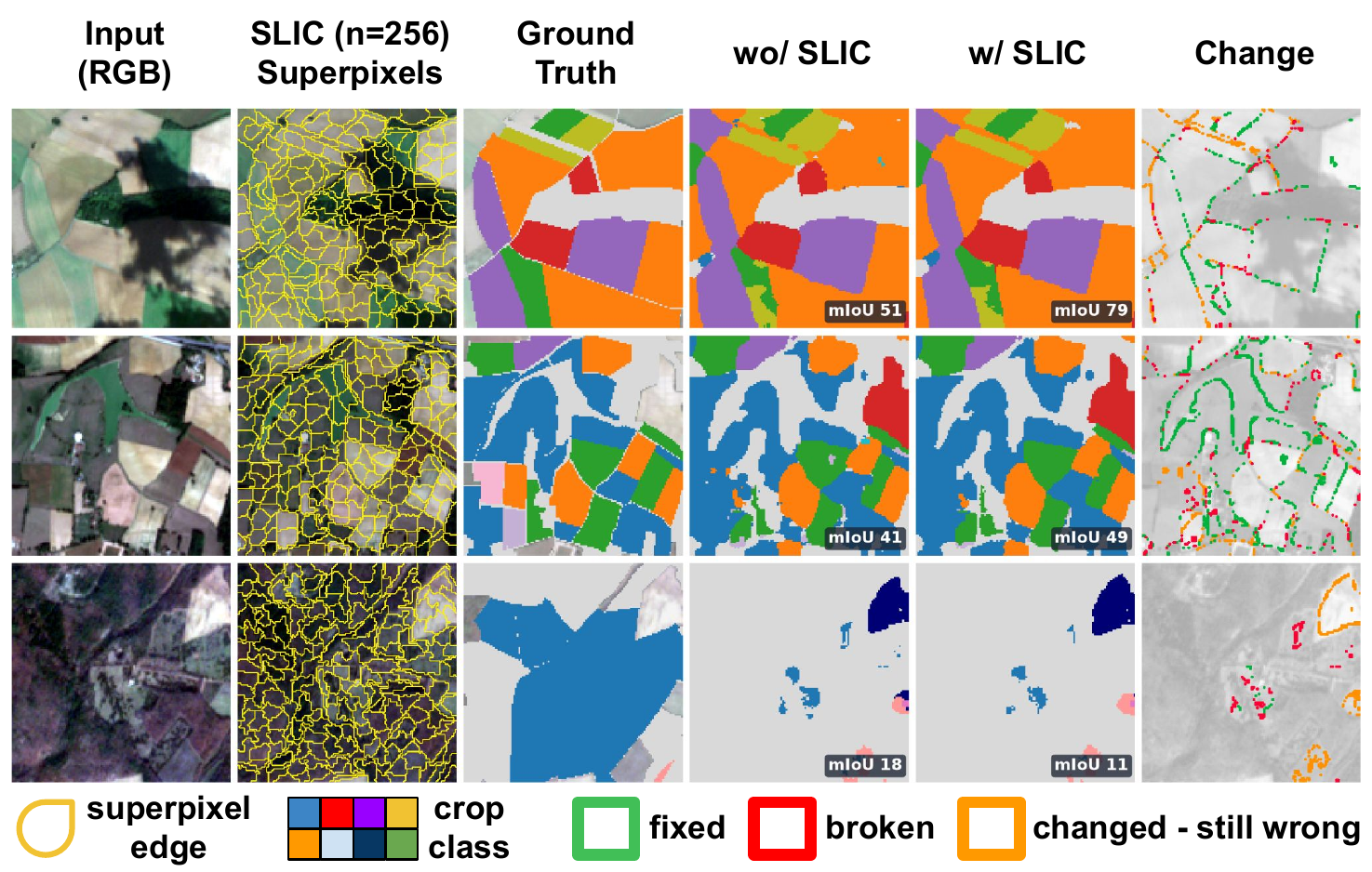}
\caption{\textbf{Effect of SLIC refinement} on PASTIS. Predictions without and with refinement are shown with their per-image mIoU (\%). The last
column marks pixels whose label changed.}    \label{fig:slic}
\end{figure*}

Tab.~\ref{tab:mados_kernel_ablations_app} reports the aggregate effect of SLIC refinement~\cite{achanta2012slic} (Eq.~\ref{eq:method_refinement}) on
MADOS, relative to the fused prediction without it. Fig.~\ref{fig:slic} complements these scores with qualitative
examples on PASTIS, using 256 superpixels computed from the guide features described in Appendix~\ref{app:method_implementation_details}.
Token-level predictions are upsampled to pixel resolution and therefore tend to blur object boundaries and small structures. Averaging the fused scores within each superpixel realigns them with image edges.
Accordingly, the changed pixels (last column) concentrate along parcel boundaries. In the first two examples, most of these changes are corrections, and per-image mIoU rises from 51\% to 79\% and from 41\% to 49\%.
The refinement propagates labels within a superpixel, so it cannot recover regions where the fused prediction is already wrong. In the third example, a large parcel is mostly missed before refinement. Averaging then removes the few correctly labeled pixels inside it, and mIoU drops from 18\% to 11\%.

\begin{table}[t]
\centering
\small
\setlength{\tabcolsep}{4pt}
\renewcommand{\arraystretch}{1.10}
\begin{tabular*}{\columnwidth}{@{\extracolsep{\fill}}lccc@{}}
\toprule
\textbf{Dataset} & \textbf{Cosine \(k\)NN} & \textbf{Kernel} & \textbf{Selected by val.} \\
\midrule
HLS Burn Scars & 84.6 & \textbf{84.9} & 84.8 \\
MADOS          & 72.0 & \textbf{72.7} & 72.0 \\
PASTIS         & \textbf{37.8} & 36.9 & \textbf{37.8} \\
xView2         & 47.2 & 47.2 & \textbf{47.4} \\
Sen1Floods11   & 79.8 & 80.0 & \textbf{80.1} \\
\midrule
\textbf{Mean mIoU} & 64.3 & 64.3 & \textbf{64.4} \\

\bottomrule
\end{tabular*}
\caption{\textbf{Cosine \(k\)NN versus Gaussian kernel prediction} with
full labels, averaged over six backbones. The first two columns always
use the same predictor. In \emph{Selected by val.}, we pick for each
dataset and backbone whichever of the two predictors scores higher on
the validation set and report its test score. \textbf{Bold} marks the
best score per row.}
\label{tab:probe-choice}
\end{table}

\section{Implementation Details}
\label{app:method_implementation_details}
This section complements Sec.~\ref{sec:method} with the implementation
details of each step: descriptor construction, Gaussianization, and
reference banks, as well as prediction and spatial decoding.
We will release the code publicly at a later date.

\subsection{Descriptor Construction}

\paragraph{Visual tokens.}
We use a frozen DINOv3 ViT-L/16. Since it expects three channels, we
feed a multispectral image as several three-band views: RGB and
false-color composites of the near-infrared, red-edge, and
shortwave-infrared bands (four views on MADOS). Each band is stretched
to \([0,1]\) between the 1st and 99th percentile of its training values
and normalized with the encoder's input statistics. From each view, we
take the final-layer patch tokens and normalize them to unit length.
Concatenating the views gives one visual descriptor for every
\(16\times16\)-pixel patch, with \(4\times1024\) dimensions on MADOS.

\paragraph{Other backbones.}
The views are only needed because DINOv3 is an RGB model. Most EO
backbones accept multispectral input directly, so each encoder
receives, in a single forward pass, the bands it was pretrained on,
normalized as its own input adapter prescribes. A band that the encoder
expects but the dataset lacks is set to its normalization mean. The
tokens are extracted as above and replace the DINOv3 descriptor, while
the rest of the pipeline is unchanged.

\paragraph{Spectral tokens.}
The second descriptor is handcrafted and needs no encoder. For the same
patches, we compute the mean, standard deviation, and mean gradient
magnitude of every band and of five spectral indices
(Tab.~\ref{tab:spectral-token-dimensions}). Let \(R,G,B,N\) denote
red, green, blue, and near-infrared reflectance, and \(S_1,S_2\) the
first and second shortwave-infrared bands. The indices are
\[
\begin{gathered}
\begin{aligned}
\mathrm{NDVI} &= \frac{N-R}{N+R+\epsilon}, &
\mathrm{NDWI} &= \frac{G-N}{G+N+\epsilon}, \\
\mathrm{NDBI} &= \frac{S_1-N}{S_1+N+\epsilon}, &
\mathrm{NBR} &= \frac{N-S_2}{N+S_2+\epsilon},
\end{aligned} \\
\mathrm{BSI} = \frac{(S_1+R)-(N+B)}{(S_1+R)+(N+B)+\epsilon},
\end{gathered}
\]
with \(\epsilon=10^{-6}\). On MADOS, 11 bands and 5 indices give a
48-dimensional descriptor.

\paragraph{Pixel descriptors.}
Visual and spectral tokens each describe a whole patch. The pixel
branch instead needs one descriptor per pixel: the band values, the
same indices except BSI, and the mean and standard deviation of every
band in a \(3\times3\) neighborhood (37 dimensions on MADOS).

\begin{table}[t]
\centering
\footnotesize
\setlength{\tabcolsep}{2pt}
\begin{tabular}{@{}llrr@{}}
\toprule
Dataset & Spectral inputs & Dim. & Footprint \\
\midrule
MADOS & 11 bands + 5 indices & 48 & \(16\times16\) \\
HLS Burn Scars & 6 bands + 5 indices & 33 & \(16\times16\) \\
Sen1Floods11 S1 & \(VV\), \(VH\), \(VV-VH\) & 8 & \(16\times16\) \\
PASTIS & (10 bands + 5 indices) \(\times 6\) & 270 & \(4\times4\) \\
xView2 & RGB at both dates & 18 & \(16\times16\) \\
EuroSAT & 13 bands + 5 indices & 54 & Whole image \\
\bottomrule
\end{tabular}
\caption{Spectral descriptors per dataset. \emph{Dim.} is the
descriptor dimension, counted over all dates. \emph{Footprint} is the
area, in input pixels, that one descriptor summarizes at each date.}
\label{tab:spectral-token-dimensions}
\end{table}

\paragraph{Temporal inputs.}
The encoder processes one image at a time, so a time series is encoded
date by date. For PASTIS, we follow PANGAEA~\cite{marsocci2026pangaea}
and use six acquisitions sampled uniformly over the series. Each frame
is upsampled by a factor of four, so that one token covers
\(4\times4\) original pixels. A PCA fitted on training data then
reduces the visual descriptor of each date to 256 components, with the
same basis for all dates.

The dates are merged at the token level: for each token, its six
per-date descriptors are concatenated in chronological order into a
single descriptor. Spectral tokens are concatenated in the same way.

Clouds hide the surface at some dates and would corrupt the result.
Since clouds are bright in the blue band, we flag a token as cloudy at
a given date when its mean blue reflectance exceeds 0.2, a simple but
effective heuristic. Before concatenation, a flagged descriptor is
replaced by linear interpolation in time between the nearest clear
dates. The pixel branch fills its band values in the same way.

xView2 has only two dates, a pre- and a post-event image. We
concatenate their visual descriptors per token, as for PASTIS. In
addition, the difference between the post- and pre-event tokens forms
a change descriptor, which is handled like the visual and spectral
ones: it predicts on its own and enters the final fusion. The pixel
branch also uses it to retrieve its reference patches.

\subsection{Gaussianization and Reference Banks}

\paragraph{Token transform.}
Each descriptor group has its own transform. It is fitted once on up
to 200{,}000 randomly sampled training tokens, without using labels,
and then applied unchanged to references and queries. It follows the
four stages of Sec.~\ref{sec:method}, with these settings:
(a) \emph{z-score}, as described there;
(b) \emph{whitening}, keeping up to \(d=768\) components for the
visual group and all components for the spectral group;
(c) \emph{marginal Gaussianization}, with \(K=1\) round, where the
empirical CDF of each coordinate is represented by a monotone
piecewise-linear map with 1{,}025 knots;
(d) \emph{radial correction}, where we map the norm of each descriptor
to the norm distribution of a standard Gaussian and keep its direction.

\paragraph{Pixel transform.}
Pixel descriptors are low-dimensional, so a lighter transform suffices.
After z-scoring, we apply a regularized PCA whitening that keeps the
components with eigenvalue \(\lambda\ge10^{-3}\) and scales them by
\((\lambda+0.1)^{-1/2}\), followed directly by radial correction. There
is no marginal step.

\paragraph{Reference banks.}
The token bank holds every training token that contains at least one
labeled pixel. Its label is the class distribution of those pixels, and
the average label over the bank gives the class prior. The pixel bank
stores the transformed pixels and labels of the same tokens. On xView2,
where background dominates, the bank is class-balanced: in every
training image, the tokens are grouped by their majority class, and
each class contributes at most a fixed number of tokens.

\paragraph{Reduced-label runs.}
Only the selected training images supply the references and the
transform fits. Input normalization, the percentile stretches of the
views, and the per-date PCA on PASTIS keep their full-training-split
statistics.

\subsection{Prediction and Spatial Decoding}

\paragraph{Token prediction.}
Both predictors let labeled references vote with their class
distributions and differ only in who votes and with what weight. The
Gaussian kernel lets every reference in the bank vote, weighted by
Eq.~(1). Its squared bandwidth is \(h^2=\rho m\), where the distance
scale
\begin{equation}
m = \operatorname{median}_{z\in\mathcal{Q}_{\mathrm{val}}}
    \min_{a\in\mathcal{B}} \lVert z-z_a\rVert_2^2
\end{equation}
is the typical squared distance from a validation descriptor in
\(\mathcal{Q}_{\mathrm{val}}\) to its nearest reference in the bank
\(\mathcal{B}\), computed separately for each group. This makes
\(\rho\) comparable across groups and datasets. Cosine \(k\)NN instead
normalizes descriptors to unit length and lets only the \(k\) most
similar references vote, with uniform or exponential weights. In both
cases, the votes are divided by \(\pi^{\alpha}\), where \(\pi\) is the
class prior, and renormalized. This counteracts class imbalance, and
\(\alpha=0\) disables it. All of these settings are selected on the
validation set (Appendix~\ref{app:selection}).

\paragraph{Pixel branch.}
The pixel branch refines the prediction inside each token. Each query
token first retrieves its 500 most similar reference tokens by cosine
similarity in the visual group (the change descriptor on xView2),
whichever token predictor is used. The labeled pixels of these tokens
form a pool that all pixels of the query token share. Each query pixel
then votes among its 50 nearest pool pixels by cosine similarity, with
exponential weights at temperature 0.05. A pixel whose pool contains no
label falls back to the class prior.

\paragraph{Fusion and refinement.}
Token scores are bilinearly upsampled to the pixel grid and combined
with the pixel-branch scores by the geometric fusion of
Sec.~\ref{sec:method}. When SLIC is selected, each pixel's fused scores
are mixed with the mean scores of its superpixel. Two datasets differ.
On PASTIS, bilinear upsampling is replaced by joint bilateral
upsampling, guided by the standardized red, near-infrared, and first
shortwave-infrared bands of all six dates. On EuroSAT, fusion acts
directly on the image-level scores, without a pixel branch or
refinement.

\paragraph{Building-level decoding.}
xView2 assigns one damage grade per building, whereas our pipeline
predicts every pixel independently. We therefore make the prediction
consistent within a building. Thresholding
\(1-P_{\mathrm{background}}\) gives a building mask, using the
validation-selected threshold when available, and each connected
component above a minimum size is treated as one building. Its pixels'
damage scores are mixed with their mean over the component. This mean
can also be combined with a building-level prediction, obtained by
pooling the visual features over the building and its surroundings and
retrieving similar annotated buildings by cosine similarity. For
efficiency, the pixel branch only runs on tokens with a fused
background probability below 0.95 and on their immediate neighbors.
Appendix~\ref{app:selection} lists the search spaces for the mixing
weights and the minimum component size. 


\section{Experimental Setup}
\label{app:experimental_setup}

\subsection{Data Splits and Label Budgets}
We follow the PANGAEA train, validation, and test splits for all five segmentation datasets~\cite{marsocci2026pangaea}. For
EuroSAT~\cite{helber2019eurosat}, we use the longitude-based spatial split~\cite{robinson2026eurosat}.
Tab.~\ref{tab:annotation-budgets} lists the partition sizes and the number of training images retained at each label budget.
The full-label setting uses all annotations in the training split. For reduced budgets, we randomly sample training images with seeds 234, 235, and 236 (Tab.~\ref{tab:annotation-budgets}). At 1\%, MADOS and
PASTIS retain only 14 images, so a random subset can miss some classes. For these two cases, we use seeds that give at least one labeled example per class: 639745, 1292002, and 1719232 for MADOS, and 236, 237, and 238 for PASTIS.

\begin{table}[h]
    \centering
    \small
    \setlength{\tabcolsep}{3pt}
    \begin{tabular}{@{}lrrrrr@{}}
        \hline
        Dataset & Train & Val. & Test & Data-Scarce & Few-Shot \\
        \hline
        MADOS           & 1,433  & 642   & 728   & 143 (10\%)   & 14 (1\%) \\
        HLS Burn Scars  & 486    & 54    & 264   & 48 (10\%)    & 4 (1\%) \\
        PASTIS          & 1,455  & 482   & 496   & 145 (10\%)   & 14 (1\%) \\
        xView2          & 13,384 & 917   & 933   & 133 (1\%)    & 13 (0.1\%) \\
        Sen1Floods11 S1 & 252    & 89    & 90    & - & - \\
        EuroSAT         & 16,200 & 5,400 & 5,400 & - & - \\
        \hline
    \end{tabular}
    \caption{Dataset splits and training images per label budget.}
    \label{tab:annotation-budgets}
\end{table}

\subsection{Hyperparameter Selection}\label{app:selection}

\paragraph{Selection procedure.}
All settings are selected on the validation set, using mIoU for
segmentation and accuracy for classification, in three steps. First, we
tune the cosine \(k\)NN and Gaussian-kernel predictors over the search
spaces in Tab.~\ref{tab:app_search}. Each predictor is tuned on its
own and separately for the visual and spectral feature groups, and the
pixel branch reuses the prior-correction exponent of its visual group.
The number of Gaussianization rounds, the retained visual dimension,
and the pixel-branch settings are fixed and not searched. Second, we
select the fusion weights, and third, the spatial refinement settings,
as described next.

\begin{table*}[t]
\centering
\small
\begin{tabular}{@{}lll@{}}
\toprule
Parameter & Visual/change group & Spectral group \\
\midrule
Cosine neighbors $k$ &
$\{10,20,50,100,200,500\}$ &
$\{10,20,50,500\}$ \\ 
Cosine weighting &
Uniform or $T\in\{0.05,0.1\}$ &
Uniform or $T\in\{0.02,0.05,0.1\}$ \\
Cosine prior exponent $\alpha_g$ &
$\{0,0.5\}$ & $\{0,0.5\}$ \\
Kernel scale $\rho$ &
$\{0.005,0.01,0.02,0.05\}$ &
$\{0.05,0.1,0.2,1\}$ \\
Kernel prior exponent $\alpha_g$ &
$\{0,0.5,1\}$ & $\{0,0.5\}$ \\
\bottomrule
\end{tabular}
\caption{Hyperparameter search spaces for the cosine \(k\)NN and
Gaussian-kernel predictors, selected on the validation set.
``Uniform'' denotes equal neighbor weights. The visual settings also
apply to the xView2 change group.}
\label{tab:app_search}
\end{table*}

\paragraph{Fusion weights.}
We search the fusion weights in two stages: a coarse grid over the
weight simplex, followed by a greedy refinement that moves weight
between pairs of branches and keeps a move only if it improves the
validation score. For single-date segmentation and EuroSAT, the grid
spacing is $0.25$ and the refinement steps are
$\{0.1,0.05,0.02,0.01\}$. PASTIS uses a grid spacing of $0.1$ and
steps $\{0.05,0.02,0.01\}$. For xView2, the pixel branch is added
afterwards: the selected token weights are scaled by
$1-w_{\mathrm{pix}}$ with $w_{\mathrm{pix}}\in\{0,0.1,0.2,0.3,0.4\}$,
followed by the same refinement as for PASTIS.

\paragraph{Superpixel refinement.}
We search the number of superpixels over $\{64,128,256,512\}$ for MADOS
and over $\{256,512,1024,2048\}$ for HLS Burn Scars and Sen1Floods11,
together with the mixing coefficient $\beta\in\{0,0.25,0.5,0.75\}$.
PASTIS uses a fixed 256 superpixels and jointly selects
$\sigma_r\in\{0.03,0.05,0.075,0.1,0.15,0.25\}$ and
$\beta\in\{0.6,0.7,0.8,0.9\}$.

\paragraph{xView2 building classifier.}
The additional building classifier for xView2 is tuned in two steps. We
first search its retrieval settings: $k\in\{1,3,5,10,20,50,100,200\}$,
uniform or exponential weights with $T\in\{0.02,0.05,0.1\}$, and prior
exponents $\{0,0.5,1\}$. These are scored by area-weighted mIoU over
the four damage classes on a fixed random third of the validation
scenes. We then select, on the validation set, how its output is
combined and decoded: the component mixing coefficient from
$\{0,0.25,0.5,0.75\}$, the building-retrieval contribution from
$\{0,0.25,0.5,0.75,1\}$, and the minimum component size from
$\{0,64\}$ pixels.

\subsection{Baseline Implementations}

\subsubsection{Trained Baselines}
For segmentation, we freeze the backbones and train Light and UPerNet decoders
on feature maps.

\paragraph{\textit{Light} decoder.}
\textit{Light} bilinearly aligns the feature maps to a common grid and concatenates them along the channel dimension. A trainable $1\times1$ convolution predicts class logits, which are bilinearly upsampled to the target resolution. On PASTIS and xView2, \textit{Light} concatenates per-date features when the encoder exposes them.

\paragraph{\textit{UPerNet} decoder.}
We use the PANGAEA implementation~\cite{marsocci2026pangaea} of
UPerNet~\cite{xiao2018unified} with 512 decoder channels. It applies pyramid pooling to the deepest feature map, fuses it top-down with the shallower maps, and bilinearly upsamples the logits to the target resolution. Inputs with several dates follow the PANGAEA protocol: the frozen backbone encodes each date independently. On PASTIS, the per-date features are aggregated over time before the decoder. PANGAEA provides a linear mapping and L-TAE (\cite{series2020lightweight}, temporal self-attention layer) for this step, and we use the linear mapping.
On xView2, the decoder is used in a Siamese fashion: the pre- and post-disaster images pass through the same encoder, and their features are concatenated at each level.

\paragraph{Training details.}
Training follows the PANGAEA protocol~\cite{marsocci2026pangaea}. We
train the decoders for 80 epochs with pixelwise cross-entropy,
excluding the dataset's ignore label, using AdamW with a learning rate of $10^{-4}$ and weight decay of $0.05$. The learning rate is reduced by a factor of ten at 60\% and 90\% of the training steps, and the batch size is typically 8. We select the checkpoint with the best validation mIoU and report its mIoU on the full test split. Sparse-label runs report the mean and sample standard deviation over three subsets.

\paragraph{EuroSAT linear probe.}
We keep the encoder frozen and add a single linear layer on top of its average-pooled features. The layer is trained for 50 epochs with batch size 32 and the AdamW schedule above. We report top-1 test accuracy for the checkpoint with the best validation accuracy.

\subsubsection{INSID3 and FROST}
\label{app:baseline_adaptations}

INSID3 \citep{cuttano2026insid3} and FROST \citep{park2026frost}
predict a binary mask for a class specified by annotated reference
images. We extend both methods to produce multiclass label maps and
to handle inputs from multiple acquisition dates. Both methods use
frozen DINOv3 ViT-L/16 features.

\paragraph{Choosing references.}
Both methods need reference images for each class. For every query
image and class, we pick up to five training images that contain this
class and look most similar to the query, using only images within the
current label budget. Similarity is the cosine similarity between
image-level descriptors, i.e., the average of the normalized patch
embeddings (one such vector per date, concatenated, for multi-date
inputs). Both methods receive the same references.

\paragraph{Multiclass prediction.}
Each method outputs one binary mask per class, and these masks must be
merged. Two cases need a rule. If several masks claim the same pixel,
we keep the most confident class: for INSID3, the one with the highest
aggregation score on MADOS and the highest prototype similarity on
PASTIS and xView2; for FROST, the one with the highest smoothed
foreground--background log density ratio. If no mask claims a pixel,
we assign the class with the most similar prototype. These rules were
selected by validation mIoU. HLS Burn Scars has a single foreground
class, so no merging is needed. On MADOS, where annotations are sparse,
we also evaluate a variant that ignores unlabeled reference tokens in
backward matching and, for FROST, in background density estimation.

\paragraph{Temporal inputs.}
Both methods expect a single image. For PASTIS (six evenly spaced
frames) and xView2 (pre- and post-disaster images), we encode each
date separately and concatenate the features of each location across
dates. Each method first applies its own feature processing per date:
positional debiasing and normalization for INSID3, and additionally
within-class shrinkage whitening for FROST. For INSID3, the similarity
between two tokens then equals the average of their per-date
similarities. FROST's bilateral propagation uses the visual features
and RGB values of all dates.

\section{Theoretical Details and Metric Definitions}
\label{app:theoretical_details}
\subsection{Descriptor Geometry in Kernel and Cosine Prediction }

The transform in Sec.~\ref{sec:method} changes both the local density of frozen
descriptors and the distribution of their norms. The two readouts
respond differently to these changes.

\paragraph{Population kernel bias.}
Let $p$ be the descriptor density and $\eta_c(z)$ the conditional
mean of the class-$c$ target. The population Nadaraya--Watson
predictor is $(G_h*(p\eta_c))/(G_h*p)$, where $G_h$ is the density of
$\mathcal{N}(0,h^2 I_d)$~\cite{nadaraya1964estimating,watson1964smooth}.
For sufficiently smooth $p$ and $p\eta_c$, expanding each convolution
as $G_h*f=f+\tfrac{h^2}{2}\Delta f+o(h^2)$ and taking their ratio
gives, at points where $p(z)>0$,
\begin{align*}
\operatorname{Bias}_c(z)
&= \tfrac{h^2}{2}\Delta\eta_c(z)
 + h^2\nabla\eta_c(z)^\top\nabla\log p(z)
 + o(h^2).
\end{align*}
The density term affects the vote when the conditional class score
varies along the density gradient. Under common bounds on the target
derivatives, the upper bound on the integrated squared leading bias
increases with the Fisher information of $p$, which an isotropic
Gaussian minimizes at fixed total variance~\cite{balestriero2025lejepa}.
This motivates the target geometry. A post-hoc transform can also
change the target derivatives, and a finite reference bank introduces
sampling error beyond the population smoothing bias.

\paragraph{Gaussian kernels and cosine voting.}
The Gaussian kernel of Eq.~(1) weights a reference $z_a$ by its squared
Euclidean distance to the query $z$,
\[
\kappa_h(z,z_a)
= \exp\!\Bigl(-\frac{\|z-z_a\|_2^2}{2h^2}\Bigr),
\]
whereas cosine voting weights it by its cosine similarity,
\[
w_T(z,z_a) = \exp\!\bigl(\cos(z,z_a)/T\bigr),
\qquad
\cos(z,z_a)=\frac{z^\top z_a}{\|z\|_2\,\|z_a\|_2},
\]
with temperature $T$. Expanding the squared distance gives
$\|z-z_a\|_2^2=\|z\|_2^2+\|z_a\|_2^2-2\,z^\top z_a$. For
unit-normalized descriptors, this reduces to
$\|z-z_a\|_2^2=2-2\cos(z,z_a)$, i.e., twice the cosine distance, and
the kernel becomes
\[
\kappa_h(z,z_a)
= e^{-1/h^2}\exp\!\bigl(\cos(z,z_a)/h^2\bigr)
= e^{-1/h^2}\,w_{h^2}(z,z_a).
\]
The kernel weight is thus the cosine weight with temperature $T=h^2$,
up to the factor $e^{-1/h^2}$. This factor is the same for every
reference, so it does not change their ranking and cancels when the
votes are normalized. The two predictors therefore give identical
predictions whenever they vote over the same references: the full-bank
Gaussian kernel equals exponentially weighted cosine voting over the
full bank, and restricting both to the same $k$ nearest references
gives exponentially weighted cosine $k$NN. Without unit normalization,
the kernel additionally depends on the descriptor norms.

\paragraph{Effect of radial correction.}
Radial correction rescales each nonzero descriptor by a positive
factor to match the target $\chi_d$ norm distribution while preserving
its direction~\cite{lyu2009radial}. Because the corrected norms
still vary, Gaussian-kernel weights depend on both norms and angles.
Cosine scores and rankings are invariant to this rescaling. Thus,
in the comparison with PCA whitening alone, any change in cosine
$k$NN predictions comes from marginal Gaussianization and
re-whitening. For Gaussian-kernel prediction, radial correction
can also contribute.

\subsection{Evaluation Metrics}
\label{sec:evaluation-metrics}

\paragraph{Mean IoU.}
We use mean intersection-over-union (mIoU) for segmentation on MADOS, HLS Burn
Scars, PASTIS, xView2, and Sen1Floods11 S1 at the ground-truth resolution. Following
PANGAEA~\cite{marsocci2026pangaea}, we pool pixel counts over the entire
evaluation split and exclude pixels with ignored ground-truth labels. For each
evaluated class $c \in \mathcal{C}$, let $\mathrm{TP}_c$, $\mathrm{FP}_c$, and
$\mathrm{FN}_c$ denote its true-positive, false-positive, and false-negative
pixel counts. We compute
\begin{equation}
    \mathrm{IoU}_c =
    \frac{\mathrm{TP}_c}{\mathrm{TP}_c + \mathrm{FP}_c + \mathrm{FN}_c},
    \qquad
    \mathrm{mIoU} = \frac{1}{|\mathcal{C}|}\sum_{c\in\mathcal{C}}\mathrm{IoU}_c .
\end{equation}
A class with an empty union contributes zero. We report mIoU as a percentage.
In the sparse-label segmentation tables, we compute test mIoU separately for
each seed and report the mean and sample standard deviation over three seeds.

\paragraph{Top-1 accuracy.}
For EuroSAT scene classification on the spatial split, we predict the class
with the highest score for each image. With $N$ evaluation images, predicted
labels $\hat{y}_i$, and ground-truth labels $y_i$, top-1 accuracy is
\begin{equation}
    \mathrm{Acc} = \frac{1}{N}\sum_{i=1}^{N}\mathbf{1}[\hat{y}_i = y_i] ,
\end{equation}
reported as a percentage.

\paragraph{Epps--Pulley discrepancy (EP).}
We measure how closely token descriptors follow an isotropic Gaussian using
the projection-based Epps--Pulley discrepancy~\cite{epps1983normality,balestriero2025lejepa}.
For descriptors $z_1,\ldots,z_n$ and random unit directions
$a_1,\ldots,a_L$, we define
\begin{equation}
    \hat{\varphi}_\ell(t) = \frac{1}{n}\sum_{j=1}^{n}
    \exp\!\left(\mathrm{i}t\,a_\ell^{\top}z_j\right) .
\end{equation}
We compare each projected empirical characteristic function with that of
$\mathcal{N}(0,1)$ and average across directions,
\begin{equation}
    \mathrm{EP} = \frac{1}{L}\sum_{\ell=1}^{L}\int_{-5}^{5}
    \bigl|\hat{\varphi}_\ell(t) - e^{-t^2/2}\bigr|^2 e^{-t^2/2}\,\mathrm{d}t .
\end{equation}
We use $L=256$ directions, up to $20{,}000$ descriptors, and trapezoidal
integration on $17$ equally spaced points. Lower values indicate closer
agreement.

\end{document}